%% file: main.tex
\documentclass[10pt,journal,compsoc]{IEEEtran}
\usepackage{amsmath,amsfonts}
\usepackage{algorithmic}
\usepackage{algorithm}
\usepackage{array}
\usepackage[caption=false,font=normalsize,labelfont=sf,textfont=sf]{subfig}
\usepackage{textcomp}
\usepackage{stfloats}
\usepackage{url}
\usepackage{verbatim}
\usepackage{graphicx}
\usepackage{cite}

\input{utils}
\begin{document}

\title{MTMamba++: Enhancing Multi-Task Dense Scene Understanding via Mamba-Based Decoders}

\author{Baijiong Lin, Weisen Jiang, Pengguang Chen, Shu Liu, and Ying-Cong Chen
\IEEEcompsocitemizethanks{\IEEEcompsocthanksitem Baijiong Lin is with Artificial Intelligence Thrust, The Hong Kong University of Science and Technology (Guangzhou), Guangzhou, China, and also with HKUST(GZ) - SmartMore Joint Lab. (E-mail: bj.lin.email@gmail.com)
\IEEEcompsocthanksitem Weisen Jiang is with Department of Computer Science and Engineering, The Hong Kong University of Science and Technology, Hong Kong, China. (E-mail: waysonkong@gmail.com)
\IEEEcompsocthanksitem Pengguang Chen and Shu Liu are with SmartMore Corporation Limited, Shenzhen, China. (E-mail: akuxcw@gmail.com, liushuhust@gmail.com)
\IEEEcompsocthanksitem Ying-Cong Chen is with Artificial Intelligence Thrust, The Hong Kong University of Science and Technology (Guangzhou), Guangzhou, China, with HKUST(GZ) - SmartMore Joint Lab, and also with Department of Computer Science and Engineering, The Hong Kong University of Science and Technology, Hong Kong, China. (E-mail: yingcongchen@ust.hk)
\IEEEcompsocthanksitem Corresponding authors: Shu Liu and Ying-Cong Chen.
}}

\markboth{Journal of \LaTeX\ Class Files,~Vol.~14, No.~8, August~2021}%
{Shell \MakeLowercase{\textit{et al.}}: A Sample Article Using IEEEtran.cls for IEEE Journals}



\IEEEtitleabstractindextext{
\begin{abstract}
Multi-task dense scene understanding, which trains a model for multiple dense prediction tasks, has a wide range of application scenarios. Capturing long-range dependency and enhancing cross-task interactions are crucial to multi-task dense prediction. In this paper, we propose MTMamba++, a novel architecture for multi-task scene understanding featuring with a Mamba-based decoder. It contains two types of core blocks: self-task Mamba (STM) block and cross-task Mamba (CTM) block. STM handles long-range dependency by leveraging state-space models, while CTM explicitly models task interactions to facilitate information exchange across tasks. 
We design two types of CTM block, namely F-CTM and S-CTM, to enhance cross-task interaction from feature and semantic perspectives, respectively.
Extensive experiments on NYUDv2, PASCAL-Context, and Cityscapes datasets demonstrate the superior performance of MTMamba++ over CNN-based, Transformer-based, and {diffusion-based} methods {while maintaining high computational efficiency}. 
The code is available at \url{https://github.com/EnVision-Research/MTMamba}.
\end{abstract}

}

\maketitle
\IEEEdisplaynontitleabstractindextext
\IEEEpeerreviewmaketitle

\IEEEraisesectionheading{\section{Introduction}} \label{sec:intro}

\IEEEPARstart{M}{ulti}-task dense scene understanding, which trains a single model to simultaneously handle multiple pixel-wise prediction tasks (e.g., semantic segmentation, depth estimation, surface normal estimation, and object boundary detection), has become increasingly important in many computer vision applications \cite{vandenhende2021multi}, such as autonomous driving \cite{liang2023multi}, healthcare \cite{hur2023genhpf}, and robotics \cite{ze2023gnfactor}. {The success of multi-task dense prediction hinges on addressing two fundamental challenges: 
\begin{enumerate*}[(i), series = tobecont, itemjoin=~]
\item modeling long-range spatial relationships to capture global context information, which is essential for pixel-wise prediction tasks;
\item enhancing cross-task interactions to facilitate knowledge sharing among different tasks, which is crucial to multi-task learning.
\end{enumerate*}}

{
Existing multi-task dense prediction approaches can be broadly categorized by their architectural design. 
CNN-based methods \cite{xu2018pad, vandenhende2020mti} employ convolutional operations in decoders for task-specific predictions but primarily capture local features, struggling with modeling long-range dependencies and global context understanding \cite{bello2019attention, li2023uni}. 
Transformer-based methods \cite{ye2022inverted, xu2023multi, ye2023invptplus, ye2023taskprompter,wang2024tsp,yang2024multi} employ attention mechanisms \cite{vaswani2017attention} to better capture global context and demonstrate improved performance. 
However, they suffer from quadratic computational complexity with respect to sequence length \cite{wu2023p2t, wang2024cross}, making them computationally prohibitive for high-resolution dense prediction tasks.}

{To address these limitations, we propose MTMamba++, a novel Mamba-based architecture that achieves effective and efficient multi-task dense scene understanding. 
MTMamba++ introduces two key components based on state space models (SSMs) \cite{gu2021combining, gu2021efficiently} in the decoder:
\begin{enumerate*}[(i), series = tobecont, itemjoin = \quad]
	\item The self-task Mamba (STM) block, inspired by \cite{liu2024vmamba}, captures global context information for each task by leveraging the long-range modeling capabilities of SSMs with linear computational complexity;
	\item The cross-task Mamba (CTM) block enables effective knowledge exchange across tasks through two variants: F-CTM for feature-level interaction and S-CTM for semantic-level interaction. The S-CTM introduces a novel cross SSM (CSSM) mechanism that models relationships between task-specific and shared feature sequences, providing more effective task interaction than simple feature fusion approaches used in F-CTM.
\end{enumerate*}}

\begin{figure*}[!t]
\centering
\includegraphics[width=\textwidth]{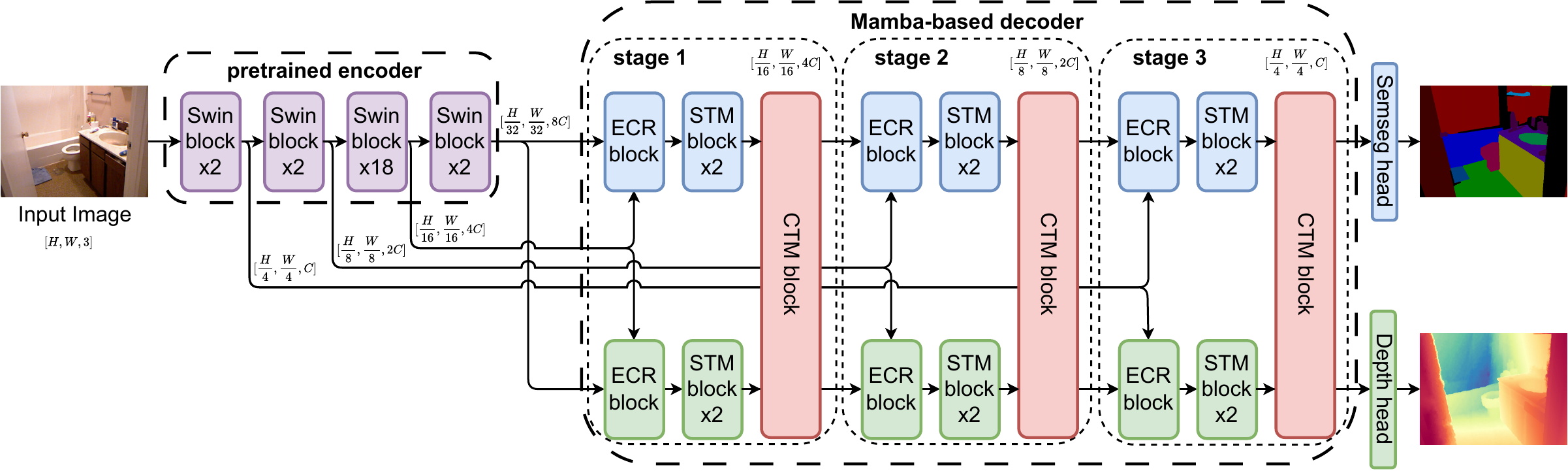}
\vspace{-0.3in}
\caption{Overview of the general architecture for MTMamba++ and its preliminary version MTMamba \cite{lin2024mtmamba}, presenting with semantic segmentation (abbreviated as ``Semseg'') and depth estimation (abbreviated as ``Depth'') tasks. The pretrained encoder (Swin-Large Transformer \cite{liu2021swin} is used here) is responsible for extracting multi-scale generic visual representations from the input RGB image. In the decoder, the ECR (expand, concatenate, and reduce) block is designed to upsample the feature maps and fuse them with high-level features derived from the encoder. Following this, the task-specific representations captured by the self-task Mamba (STM) blocks are further refined in the cross-task Mamba (CTM) block. This process ensures that each task benefits from the comprehensive feature set provided by the shared and task-specific components. Each task has its own head to generate the final predictions. 
We develop two types of CTM blocks and prediction heads, respectively. MTMamba++ and MTMamba utilize different CTM blocks and prediction heads as their default configurations.
The details of each part are comprehensively introduced in Section \ref{sec:method}.}
\label{fig:overall_arch}
\end{figure*}

{As the overall framework shown in Figure \ref{fig:overall_arch}, MTMamba++ features a three-stage Mamba-based decoder that progressively refines multi-task predictions.
Each stage contains an ECR (expand, concatenate, and reduce) block that upscales features and fuses them with encoder features, followed by STM and CTM blocks for task-specific learning and cross-task interaction. This design effectively captures long-range dependencies and enhances cross-task interaction while maintaining computational efficiency.}

{
We evaluate MTMamba++ on three standard multi-task dense prediction benchmark datasets, namely NYUDv2 \cite{silberman2012indoor}, PASCAL-Context \cite{chen2014detect}, and Cityscapes \cite{CordtsORREBFRS16}. Quantitative results demonstrate that MTMamba++ significantly surpasses previous methods, including CNN-based, Transformer-based, and diffusion-based appoarch. Moreover, comprehensive efficiency analysis shows that MTMamba++ achieves state-of-the-art performance while maintaining high computational efficiency. Notably, our experiments demonstrate that SSM-based architectures are more effective and efficient than attention-based for multi-task dense prediction tasks.
Additionally, qualitative studies show that MTMamba++ generates superior visual results with greater accuracy in detail, sharper boundaries, and more accurate detection in small objects compared to existing approaches.}

In summary, the main contributions of this paper are four-fold:
\begin{itemize}
\item We propose MTMamba++, a novel multi-task architecture based on state space models (SSMs) for multi-task dense scene understanding. It contains a novel Mamba-based decoder, effectively modeling long-range spatial relationships and achieving cross-task correlation;
\item In the decoder, we design two types of cross-task Mamba (CTM) blocks, namely F-CTM and S-CTM, to enhance cross-task interaction from feature and semantic perspectives, respectively;
\item In the S-CTM block, we propose a novel cross SSM (CSSM) to model the relationship between two sequences based on the SSM mechanism;
\item We evaluate MTMamba++ on three benchmark datasets, including NYUDv2, PASCAL-Context, and Cityscapes. Quantitative results demonstrate the superiority of MTMamba++ on multi-task dense prediction over previous methods {while maintaining high computational efficiency}. Qualitative evaluations show that MTMamba++ generates precise predictions.
\end{itemize}

A preliminary version of this work appeared in a conference paper \cite{lin2024mtmamba}. Compared with the previous conference version, we propose a novel cross SSM (CSSM) mechanism that enables capturing the relationship between two sequences based on the SSM mechanism. By leveraging CSSM, we design a novel cross-task Mamba (CTM) block (i.e., S-CTM) to better achieve cross-task interaction. We also introduce a more effective and lightweight prediction head. Based on these innovations, MTMamba++ largely outperforms MTMamba \cite{lin2024mtmamba}. Moreover, we extend our experiments to investigate the effectiveness of MTMamba++ on a new multi-task scene understanding benchmark dataset, i.e., Cityscapes \cite{CordtsORREBFRS16}. We also provide more results and analysis to understand the proposed MTMamba++ model.

The rest of the paper is organized as follows. In Section \ref{sec:related_work}, we review some related works. In Section \ref{sec:method}, we present a detailed description of the various modules within our proposed MTMamba++ model. In Section \ref{sec:exp}, we quantitatively and qualitatively evaluate the proposed MTMamba++ model on three benchmark datasets (NYUDv2 \cite{silberman2012indoor}, PASCAL-Context \cite{chen2014detect}, and Cityscapes \cite{CordtsORREBFRS16}). Finally, we make conclusions in Section \ref{sec:conclusion}.

\section{Related Works} \label{sec:related_work}

\subsection{Multi-Task Learning}

Multi-task learning (MTL) is a learning paradigm that aims to jointly learn multiple related tasks using a single model \cite{zhang2021survey,chen2025modl}. 
Current MTL research mainly focuses on multi-objective optimization \cite{linreasonable, lin2023scale, ye2021multi, ye2024first, liu2021conflict, ye2024moml} and network architecture design \cite{ye2022inverted, ye2023invptplus, ye2023taskprompter, xu2018pad, vandenhende2020mti, xu2023multi,wang2024tsp,yang2024multi,yang2025multi,shi2023deep}. 
In multi-task visual scene understanding, most existing works focus on designing architecture \cite{vandenhende2021multi}, especially developing specific modules in the decoder to facilitate knowledge exchange among different tasks. 
For instance, based on CNN, Xu et al. \cite{xu2018pad} introduce PAD-Net, which integrates an effective multi-modal distillation module aimed at enhancing information exchange among various tasks within the decoder. 
MTI-Net \cite{vandenhende2020mti} is a complex multi-scale and multi-task CNN architecture that facilitates information distillation across various feature scales.
As the convolution operation only captures local features \cite{bello2019attention}, recent approaches \cite{ye2022inverted, ye2023invptplus, ye2023taskprompter, xu2023multi,wang2024tsp,yang2024multi} develop Transformer-based decoders to grasp global context by attention mechanism \cite{vaswani2017attention}. 
For example, InvPT \cite{ye2022inverted} is a Transformer-based multi-task architecture that employs an effective UP-Transformer block for multi-task feature interaction at different feature scales. 
MQTransformer \cite{xu2023multi} uses a cross-task query attention module in the decoder to enable effective task association and information communication.

These works demonstrate the significance of long-range dependency modeling and the enhancement of cross-task correlation for multi-task dense scene understanding. 
Different from existing methods, we propose a novel multi-task architecture derived from the SSM mechanism \cite{gu2023mamba} to capture global information better and promote cross-task interaction. 

\subsection{State Space Models}
State space models (SSMs) are a mathematical framework for characterizing dynamic systems, capturing the dynamics of input-output relationships via a hidden state. 
SSMs have found broad applications in various fields such as reinforcement learning \cite{hafner2019dream}, computational neuroscience \cite{friston2003dynamic}, and linear dynamical systems \cite{hespanha2018linear}. 
Recently, SSMs have emerged as an alternative mechanism to model long-range dependencies in a manner that maintains linear complexity with respect to sequence length.
Compared with the convolution operation, which excels at capturing local dependence, SSMs exhibit enhanced capabilities for modeling long sequences. 
Moreover, in contrast to attention mechanism \cite{vaswani2017attention}, which incurs quadratic computational costs with respect to sequence length \cite{wu2023p2t, wang2024cross}, SSMs are more computation- and memory-efficient.

To improve the expressivity and efficiency of SSMs, many different structures have been proposed. 
Gu et al. \cite{gu2021efficiently} propose structured state space models (S4) to enhance computational efficiency by decomposing the state matrix into low-rank and normal matrices. 
Many follow-up works attempt to improve the effectiveness of S4.
For instance, Fu et al. \cite{fu2023hungry} propose a new SSM layer called H3 to reduce the performance gap between SSM-based networks and Transformers in language modeling. 
Mehta et al. \cite{mehta2023long} introduce a gated state space layer leveraging gated units to enhance the models' expressive capacity.

Recently, Gu and Dao \cite{gu2023mamba} propose a new SSM-based architecture termed Mamba, which incorporates a new SSM called S6. This SSM is an input-dependent selection mechanism derived from S4. 
Mamba has demonstrated superior performance over Transformers on various benchmarks, such as language modeling \cite{gu2023mamba,grazzi2024mamba,wang2024mambabyte}, graph reasoning \cite{wang2024graph,behrouz2024graph}, medical image analysis \cite{ma2024u,xing2024segmamba}, and image classification \cite{liu2024vmamba, zhu2024vision}.
Different from existing research efforts on Mamba, which mainly focus on single-task settings, in this paper, we consider a more challenging multi-task setting and propose a novel cross-task Mamba module to capture inter-task dependence.

\section{Methodology} \label{sec:method}

In this section, we begin with the foundational knowledge of state space models (Section \ref{sec:ssm}) and provide an overview of the proposed MTMamba++ in Section \ref{sec:overview}. 
Next, we delve into a detailed exploration of each component in the decoder of MTMamba++, including the encoder in Section \ref{sec:encoder}, three types of block in the decoder (i.e., the ECR block in Section \ref{sec:ecr}, the STM block in Section \ref{sec:stm}, and the CTM block in Section \ref{sec:ctm}), and the prediction head in Section \ref{sec:head}.

\subsection{Preliminaries} \label{sec:ssm}

SSMs \cite{gu2021combining, gu2021efficiently, gu2023mamba}, 
derived from the linear systems theory \cite{hespanha2018linear}, map an input sequence $x(t)\in \bR$ to an output sequence $y(t)\in \bR$ though a hidden state $\vh\in\bR^N$ using a linear ordinary differential equation:
\begin{align}
\vh'(t) &= \vA \vh(t) + \vB x(t), \label{eq:ssm-1}\\ 
y(t) &= \vC^\top \vh(t) + D x(t), \label{eq:ssm-2}   
\end{align}
where $\vA \in \bR^{N\times N}$ is the state transition matrix,
$\vB \in \bR^{N}$ and $\vC \in \bR^{N}$ are projection matrices, 
and $D \in \bR$ is the skip connection. Equation \eqref{eq:ssm-1} defines the evolution of the hidden state $\vh(t)$, while Equation \eqref{eq:ssm-2} specifies that the output is derived from a linear transformation of the hidden state $\vh(t)$ combined with a skip connection from the input $x(t)$.

Given that continuous-time systems are not compatible with digital computers and the discrete nature of real-world data, a discretization process is essential. 
This process approximates the continuous-time system with a discrete-time one. 
Let $\Delta\in \bR$ be a discrete-time step size.
Equations \eqref{eq:ssm-1} and \eqref{eq:ssm-2} are discretized as 
\begin{align}
\vh_t &= \bar{\vA}\vh_{t-1} + \bar{\vB} x_t, \label{eq:ssm-d1}  \\ 
y_t &= \bar{\vC}^\top \vh_t+ D x_t, 
\label{eq:ssm-d2}
\end{align}
where $x_t = x(\Delta t)$, and
\begin{align}
\bar{\vA} &= \exp(\Delta \vA), \nonumber \\
\bar{\vB} &= (\Delta \vA)^{-1}(\exp(\Delta \vA) - \vI) \cdot \Delta \vB \approx \Delta \vB, \nonumber \\
\bar{\vC} &= \vC. 
\label{eq:discrete}
\end{align}

S4 \cite{gu2021efficiently} treats $\vA, \vB, \vC$, and $\Delta$ as trainable parameters and optimizes them by gradient descent. However, these parameters do not explicitly depend on the input sequence, which can lead to suboptimal extraction of contextual information. To address this limitation, Mamba \cite{gu2023mamba} introduces a new SSM, namely S6. As illustrated in Figure \ref{fig:ssm}(a), it incorporates an input-dependent selection mechanism that enhances the system's ability to discern and select relevant information contingent upon the input sequence. Specifically, $\vB, \vC$, and $\Delta$ are defined as functions of the input $\vx\in \bR^{B\times L \times C}$. Following the computation of these parameters, $\bar{\vA}, \bar{\vB}$, and $\bar{\vC}$ are calculated via Equation \eqref{eq:discrete}. Subsequently, the output sequence $\vy\in\bR^{B\times L \times C}$ is computed by Equations \eqref{eq:ssm-d1} and \eqref{eq:ssm-d2}, thereby improving the contextual information extraction.
Without specific instructions, in this paper, S6 \cite{gu2023mamba} is used in the SSM mechanism.

\begin{figure*}[!t]
\centering
\includegraphics[width=\textwidth]{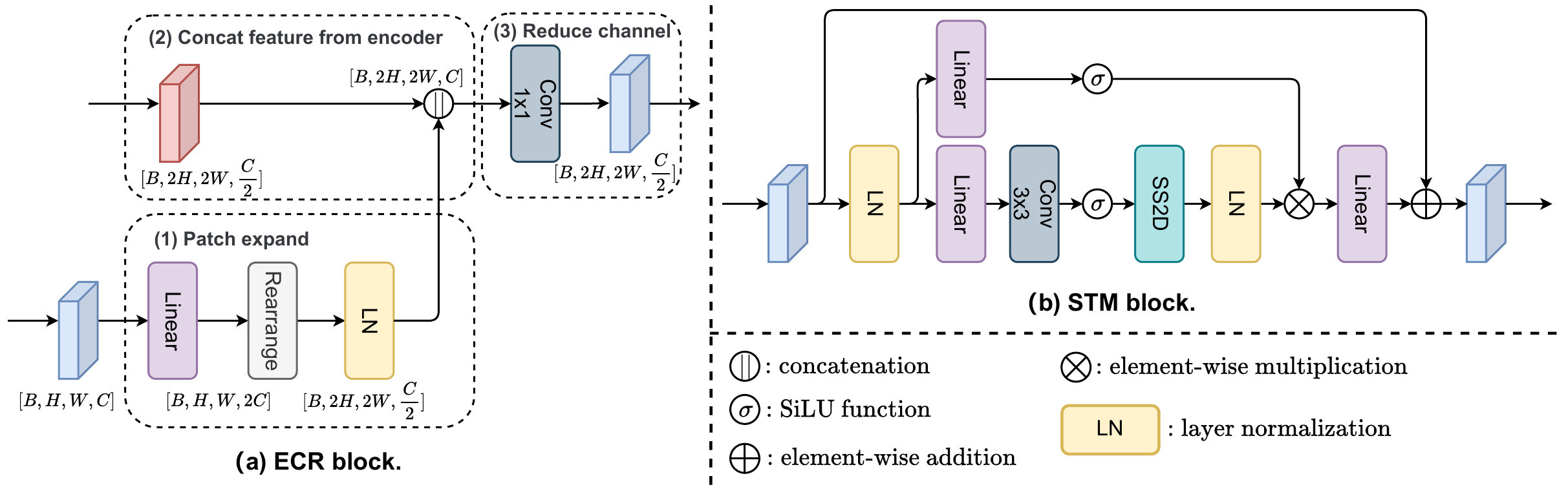}
\vspace{-0.25in}
\caption{\textbf{(a)} Illustration of the ECR (expand, concatenate, and reduce) block. It is responsible for upsampling the task feature and fusing it with the multi-scale feature from the encoder. More details are provided in Section \ref{sec:ecr}. \textbf{(b)} Overview of the self-task Mamba (STM) block, which is responsible for learning discriminant features for each task. Its core module SS2D is derived from \cite{liu2024vmamba}. As shown in Figure \ref{fig:ssm}(b), SS2D extends 1D SSM operation (introduced in Section \ref{sec:ssm}) to process 2D images. More details about STM are put in Section \ref{sec:stm}.}
\label{fig:block}
\end{figure*}

\subsection{Overall Architecture} \label{sec:overview}

An overview of MTMamba++ is illustrated in Figure \ref{fig:overall_arch}. 
It contains three components: an off-the-shelf encoder, a Mamba-based decoder, and task-specific prediction heads. 
Specifically, the encoder is shared across all tasks and plays a pivotal role in extracting multi-scale generic visual representations from the input image. 
The decoder consists of three stages, each of which progressively expands the spatial dimensions of the feature maps. This expansion is crucial for dense prediction tasks, as the resolution of the feature maps directly impacts the accuracy of the pixel-level predictions \cite{ye2022inverted}. Each decoder stage is equipped with the ECR block designed to upsample the feature and integrate it with high-level features derived from the encoder. Following this, the STM block is employed to capture the long-range spatial relationship for each task. Additionally, the CTM block facilitates feature enhancement for each task by promoting knowledge exchange across different tasks. We design two types of CTM block, namely F-CTM and S-CTM, as introduced in Section \ref{sec:ctm}. In the end, a prediction head is used to generate the final prediction for each task. We introduce two types of head, called DenseHead and LiteHead, as described in Section \ref{sec:head}.

MTMamba++ and our preliminary version MTMamba \cite{lin2024mtmamba} have a similar architecture.
The default configuration for MTMamba++ utilizes the S-CTM block and LiteHead, while the default configuration for MTMamba employs the F-CTM block and DenseHead.

\subsection{Encoder} \label{sec:encoder}

The encoder in MTMamba++ is shared across different tasks and is designed to learn generic multi-scale visual features from the input RGB image. 
As an example, we consider the Swin Transformer \cite{liu2021swin}, which segments the input image into non-overlapping patches. 
Each patch is treated as a token, and its feature representation is a concatenation of the raw RGB pixel values. 
After patch segmentation, a linear layer is applied to project the raw token into a $C$-dimensional feature embedding. 
The projected tokens then sequentially pass through four stages of the encoder. 
Each stage comprises multiple Swin Transformer blocks and a patch merging layer. 
The patch merging layer is specifically utilized to downsample the spatial dimensions by a factor of $2\times$ and expand the channel numbers by a factor of $2\times$, while the Swin Transformer blocks are dedicated to learning and refining the feature representations. 
Finally, for an input image with dimensions $H \times W \times 3$, where $H$ and $W$ denote the height and width, the encoder generates hierarchical feature representations at four different scales, i.e., $\frac{H}{4} \times \frac{W}{4} \times C$, $\frac{H}{8} \times \frac{W}{8} \times 2C$, $\frac{H}{16} \times \frac{W}{16} \times 4C$, and $\frac{H}{32} \times \frac{W}{32} \times 8C$.

\begin{figure*}[!t]
\centering
\includegraphics[width=\textwidth]{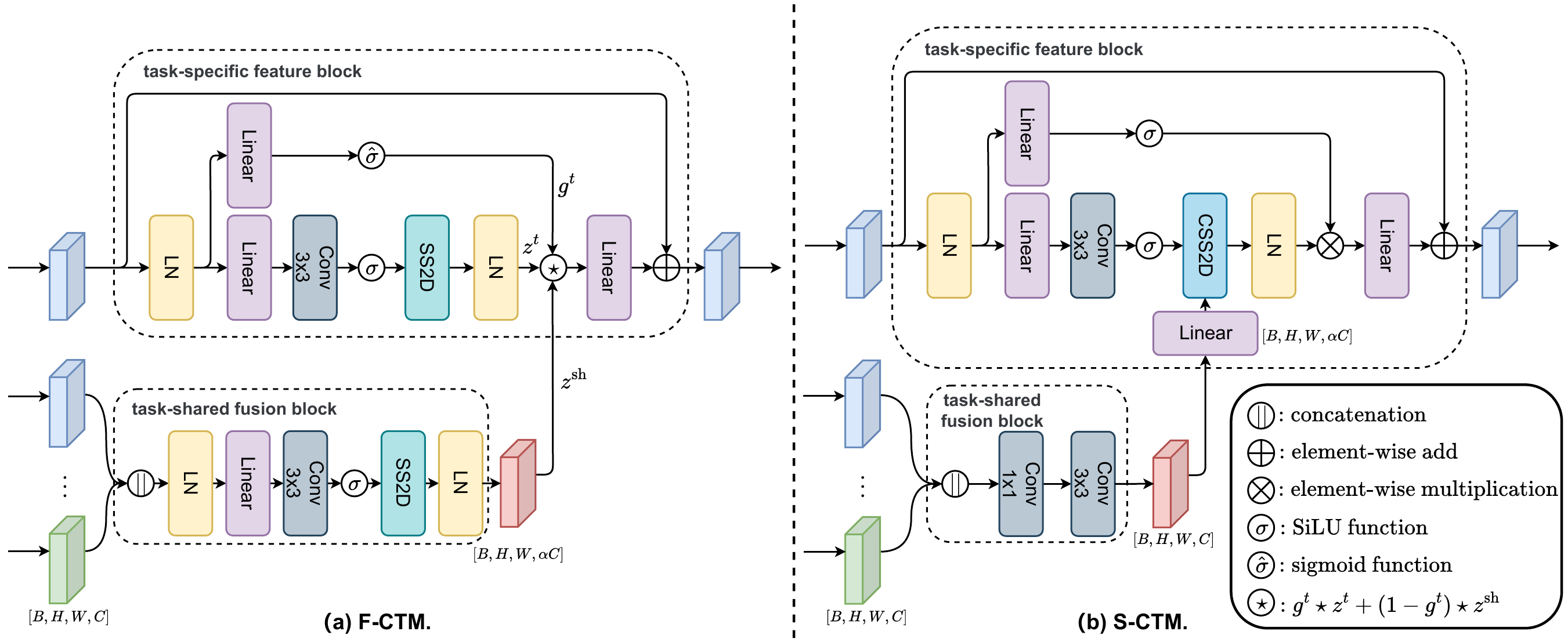}
\vspace{-0.25in}
\caption{Illustration of two types of cross-task Mamba (CTM) block. 
\textbf{(a)} F-CTM contains a task-shared fusion block for generating a global representation $\vz^{\text{sh}}$ and $T$ task-specific feature blocks (only one is illustrated) for obtaining each task's feature ${\vz}^t$. Each task's output is the aggregation of task-specific feature ${\vz}^t$ and global feature ${\vz}^\text{sh}$ weighted by a task-specific gate $\vg^t$. More details about F-CTM are provided in Section \ref{sec:F-CTM}.
\textbf{(b)} Similar to F-CTM, S-CTM generates a global feature by a fusion block and processes each task's feature with a task-specific block (only one is illustrated). Differently, S-CTM achieves semantic-aware cross-task interaction in the cross SS2D (CSS2D) module, which is shown in Figure \ref{fig:ssm}(d). More details about S-CTM and CSS2D are provided in Section \ref{sec:S-CTM}.}
\label{fig:ctm}
\end{figure*}

\subsection{ECR Block} \label{sec:ecr}

The ECR (expand, concatenate, and reduce) block is responsible for upsampling the feature and aggregating it with the encoder's feature. As illustrated in Figure \ref{fig:block}(a), it contains three steps. For an input feature, ECR block first $2\times$ upsamples the feature resolution and $2\times$ reduces the channel number by a linear layer and the rearrange operation. Then, the feature is fused with the high-level feature from the encoder through skip connections. Fusing these features is crucial for compensating the loss of spatial information that occurs due to downsampling in the encoder. Finally, a $1\times1$ convolutional layer is used to reduce the channel number.
Consequently, the ECR block facilitates the efficient recovery of high-resolution details, which is essential for dense prediction tasks that require precise spatial information.

\subsection{STM Block} \label{sec:stm}

The self-task Mamba (STM) block is responsible for learning task-specific features. As illustrated in Figure \ref{fig:block}(b), its core module is the 2D-selective-scan (SS2D) module, which is derived from \cite{liu2024vmamba}. The SS2D module is designed to address the limitations of applying 1D SSMs (as discussed in Section \ref{sec:ssm}) to process 2D image data. As depicted in Figure \ref{fig:ssm}(b), it unfolds the feature map along four distinct directions, creating four unique feature sequences, each of which is then processed by an SSM. The outputs from four SSMs are subsequently added and reshaped to form a comprehensive 2D feature map.

For an input feature, the STM block operates through several stages: it first employs a linear layer to expand the channel number by a controllable expansion factor $\alpha$.
A convolutional layer with a SiLU activation function is used to extract local features. 
The SS2D operation models the long-range dependencies within the feature map.
An input-dependent gating mechanism is integrated to adaptively select the most salient representations derived from the SS2D process.
Finally, another linear layer is applied to reduce the expanded channel number, yielding the output feature.
Therefore, the STM block effectively captures both local and global spatial information, which is essential for the accurate learning of task-specific features in dense scene understanding tasks.

\subsection{CTM Block} \label{sec:ctm}

While the STM block excels at learning distinctive representations for individual tasks, it fails to establish inter-task connections, which are essential for enhancing the performance of MTL. 
To address this limitation, we propose the novel cross-task Mamba (CTM) block, depicted in Figure \ref{fig:ctm}, which facilitates information exchange across various tasks.
We develop two types of CTM blocks, called F-CTM and S-CTM, from different perspectives to achieve cross-task interaction.

\subsubsection{F-CTM: Feature-Level Interaction} \label{sec:F-CTM}
As shown in Figure \ref{fig:ctm}(a), F-CTM comprises a task-shared fusion block and $T$ task-specific feature blocks, where $T$ is the number of tasks. It inputs $T$ features and outputs $T$ features. For each task, the input features have a channel dimension of $C$. 

The task-shared fusion block first concatenates all task features, resulting in a concatenated feature with a channel dimension of $TC$. This concatenated feature is then fed into a linear layer to transform the channel dimension from $TC$ to $\alpha C$, aligning it with the dimensions of the task-specific features from the task-specific feature blocks, where $\alpha$ is the expansion factor introduced in Section \ref{sec:stm}. The transformed feature is subsequently processed through a sequence of operations ``Conv - SiLU - SS2D" to learn a global representation ${\vz}^\text{sh}$, which contains information from all tasks. 

In the task-specific feature block, each task independently processes its own feature representation ${\vz}^t$ through its own sequence of operations ``Linear - Conv - SiLU - SS2D". Then, we use a task-specific and input-dependent gate $\vg^t$ to aggregate the task-specific representation ${\vz}^t$ and the global representation ${\vz}^\text{sh}$ as $\vg^t \times \vz^t + (1 - \vg^t) \times {\vz}^\text{sh}$. 

Hence, F-CTM allows each task to adaptively integrate the cross-task representation with its own feature, promoting information sharing and interaction among tasks. The use of input-dependent gates ensures that each task can selectively emphasize either its own feature or the shared global representation based on the input data, thereby enhancing the model's ability to learn discriminative features in a multi-task learning context.

\subsubsection{S-CTM: Semantic-Aware Interaction} \label{sec:S-CTM}

While feature fusion in F-CTM is an effective way to interact with information, it may not be sufficient to capture all the complex relationships across different tasks, especially in multi-task scene understanding where the interactions between multiple pixel-level dense prediction tasks are highly dynamic and context-dependent. Thus, we propose S-CTM to achieve semantic-aware interaction. 

As shown in Figure \ref{fig:ctm}(b), S-CTM contains a task-shared fusion block and $T$ task-specific feature blocks. The fusion block first concatenates all task features and then passes the concatenated feature through two convolution layers to generate the global representation, which contains knowledge across all tasks. The task-specific feature block in S-CTM is adapted from the STM block by replacing the SS2D with a novel cross SS2D (CSS2D). The additional input of CSS2D is from the task-shared fusion block. 

\begin{figure*}[!t]
\centering
\includegraphics[width=\textwidth]{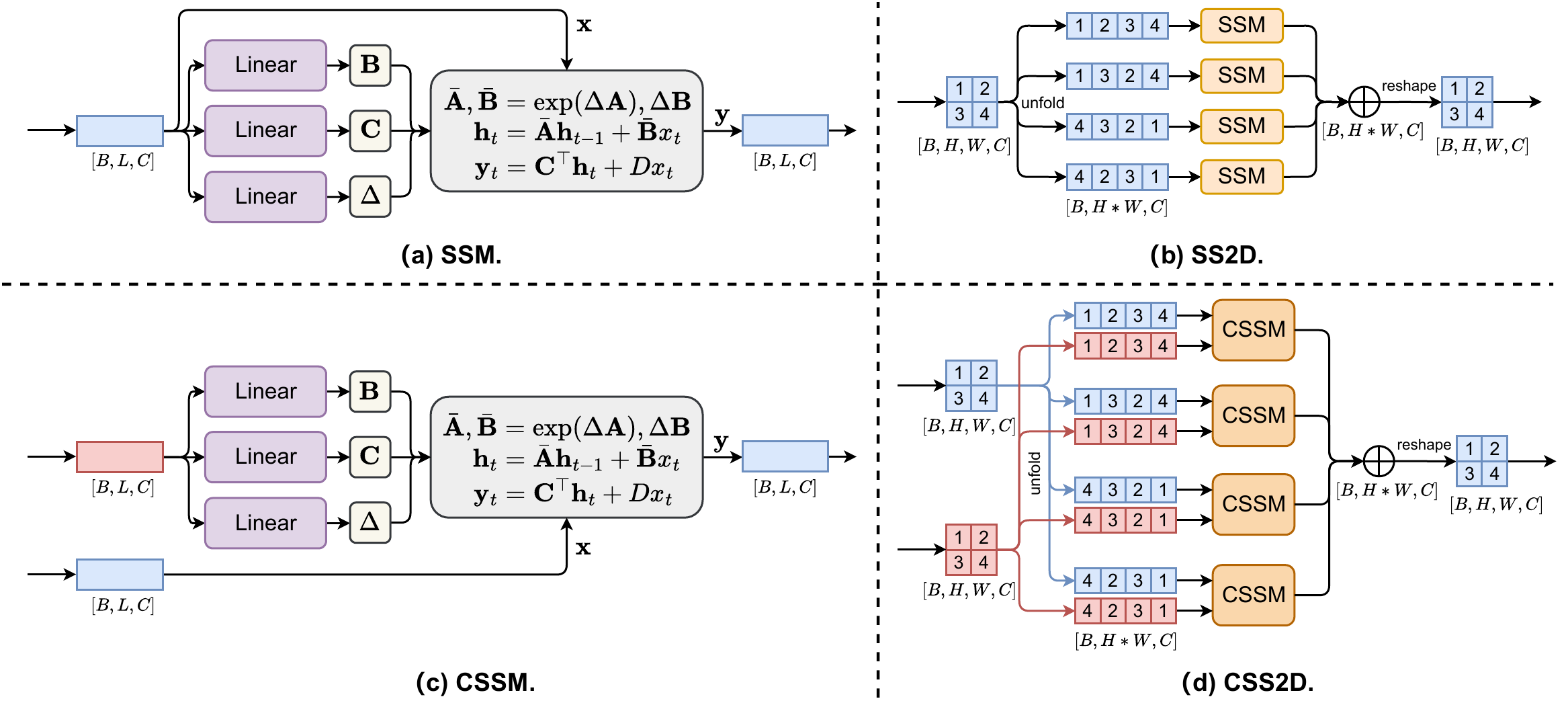}
\vspace{-0.3in}
\caption{\textbf{(a)} Illustration of SSM. Given an input sequence, SSM first computes the input-dependent parameters (i.e., $\vB, \vC$, and $\Delta$) and then calculates the output by querying the input through Equations \eqref{eq:ssm-d1} and \eqref{eq:ssm-d2}. More details about SSM are provided in Section \ref{sec:ssm}. 
\textbf{(b)} Overview of SS2D from \cite{liu2024vmamba}, which extends 1D SSMs to process 2D images. It unfolds the 2D feature map along four directions, generating four different feature sequences, each of which is then fed into an SSM. The four outputs are aggregated and folded to the 2D feature.
\textbf{(c)} Illustration of the proposed cross SSM (CSSM), which enables modeling the relationships between two input sequences based on the SSM mechanism. In CSSM, one input sequence is used to compute (i.e., $\vB, \vC$, and $\Delta$) and the other input is considered as the query. The output of CSSM is computed via Equations \eqref{eq:ssm-d1} and \eqref{eq:ssm-d2}. More details about CSSM are provided in Section \ref{sec:ctm}.   
\textbf{(d)} Overview of the proposed cross SS2D (CSS2D). It inputs two 2D feature maps, scans them along four directions to generate four pairs of feature sequences, and then passes each pair through a CSSM. The outputs of CSSMs are subsequently added and reshaped to form a final 2D output feature. The details of CSS2D are put in Section \ref{sec:S-CTM}.
}
\label{fig:ssm}
\end{figure*}

As discussed in Section \ref{sec:ssm}, SSM only models the internal relationship within a single input sequence, but it does not capture the interactions between two different sequences. 
To address this limitation, we propose the cross SSM (CSSM) to model the relationship between the task-specific feature sequence (blue) and the task-shared feature sequence (red),
as illustrated in Figure \ref{fig:ssm}(c). CSSM receives two sequences as input and outputs one sequence. 
The task-shared feature sequence is used to generate the SSMs parameters (i.e., $\vB, \vC$, and $\Delta$), and the task-specific feature sequence is considered as the query input $\vx$. 
The output is computed via Equations \eqref{eq:ssm-d1} and \eqref{eq:ssm-d2}. Consequently, by leveraging the SSM mechanism, CSSM can capture the interactions between two input sequences at the semantic level. 
Furthermore, we extend SS2D as CSS2D, as shown in Figure \ref{fig:ssm}(d). 
This module takes two 2D input features, expands them along four directions to generate four pairs of feature sequences, and feeds each pair into a CSSM. 
The outputs from these sequences are subsequently aggregated and reshaped to form a 2D output feature.

Therefore, compared with F-CTM, S-CTM can better learn context-aware relationships because of the CSSM mechanism. 
CSSM can explicitly and effectively model long-range spatial relationships within two sequences, allowing S-CTM to understand the interactions between task-specific features and the global representation, which is critical for multi-task learning scenarios. In contrast, the feature fusion in F-CTM makes it difficult to capture the complex dependencies inherent across tasks.

\subsection{Prediction Head} \label{sec:head}

As shown in Figure \ref{fig:overall_arch}, after the decoder, the size of task-specific feature is $\frac{H}{4}\times \frac{W}{4}\times C$. Each task has its own prediction head to generate its final prediction. We introduce two types of prediction heads as follows.

\subsubsection{DenseHead} 
DenseHead is inspired by \cite{cao2022swin} and is used in our preliminary version MTMamba \cite{lin2024mtmamba}. Specifically, each head contains a patch expand operation and a final linear layer. The patch expanding operation, similar to the one in the ECR block (as shown in Figure \ref{fig:block}(a)), performs $4\times$ upsampling to restore the resolution of the feature maps to the original input resolution $H \times W$. The final linear layer is used to project the feature channels to the task's output dimensions and output the final pixel-wise prediction.

\subsubsection{LiteHead} 
In DenseHead, upsampling is performed first, which can lead to a significant computational cost. Hence, we introduce a more simple, lightweight, and effective head architecture, called LiteHead. Specifically, it consists of a $3\times 3$ convolutional layer, followed by a batch normalization layer, a ReLU activation function, and a final linear layer that projects the feature channels onto the task's output dimensions. Subsequently, the feature is simply interpolated to align with the input resolution and then used as the output. Thus, LiteHead is much more computationally efficient than DenseHead. Note that since each task has its own head, the overall computational cost reduction is linearly related to the number of tasks.

\section{Experiments} \label{sec:exp}
In this section, we conduct extensive experiments to evaluate the proposed MTMamba++ in multi-task dense scene understanding. 

\subsection{Experimental Setups}
\subsubsection{Datasets} 
Following \cite{ye2022inverted, ye2023invptplus, ye2023taskprompter}, we conduct experiments on three multi-task dense prediction benchmark datasets: 
\begin{enumerate*}[(i), series = tobecont, itemjoin = \quad]
\item NYUDv2 \cite{silberman2012indoor} contains a number of indoor scenes, including 795 training images and 654 testing images. It consists of four tasks: $40$-class semantic segmentation (Semseg), monocular depth estimation (Depth), surface normal estimation (Normal), and object boundary detection (Boundary). 
\item PASCAL-Context \cite{chen2014detect}, originated from the PASCAL dataset \cite{everingham2010pascal}, includes both indoor and outdoor scenes and provides pixel-wise labels for tasks like semantic segmentation, human parsing (Parsing), and object boundary detection, with additional labels for surface normal estimation and saliency detection tasks generated by \cite{maninis2019attentive}. It contains 4,998 training images and 5,105 testing images.
\item Cityscapes \cite{CordtsORREBFRS16} is an urban scene understanding dataset. It has two tasks (19-class semantic segmentation and disparity estimation) with 2,975 training and 500 testing images.
\end{enumerate*}

\subsubsection{Implementation Details} 
We use the Swin-Large Transformer \cite{liu2021swin} pretrained on the ImageNet-22K dataset \cite{deng2009imagenet} as the encoder. The expansion factor $\alpha$ is set to $2$ in both STM and CTM blocks. Following \cite{ye2022inverted, ye2023invptplus, ye2023taskprompter}, we resize the input images of NYUDv2, PASCAL-Context, and Cityscapes datasets as $448\times 576$, $512\times 512$, and $512 \times 1024$, respectively, and use the same data augmentations including random color jittering, random cropping, random scaling, and random horizontal flipping. The $\ell_1$ loss is used for depth estimation and surface normal estimation tasks, while the cross-entropy loss is for other tasks. The proposed model is trained with a batch size of 4 for 40,000 iterations. The AdamW optimizer \cite{loshchilov2018decoupled} with a weight decay of $1\times10^{-6}$ and the polynomial learning rate scheduler are used for all three datasets. The learning rate is set to $2\times10^{-5}$, $8\times10^{-5}$, and $1\times10^{-4}$ for NYUDv2, PASCAL-Context, and Cityscapes datasets, respectively.

\begin{table*}[!t]
\centering
\caption{Comparison with state-of-the-art methods on NYUDv2 (\textbf{left}) and PASCAL-Context (\textbf{right}) datasets. $\uparrow (\downarrow)$ indicates that a higher (lower) result corresponds to better performance.  The best and second best results are highlighted in \textbf{bold} and \underline{underline}, respectively.}
\vspace{-0.1in}
\resizebox{0.45\linewidth}{!}{
\begin{tabular}{lcccc}
\toprule
\multirow{2}{*}{\textbf{Method}} & \textbf{Semseg} & \textbf{Depth} & \textbf{Normal} & \textbf{Boundary}\\
& mIoU$\uparrow$ & RMSE$\downarrow$ & mErr$\downarrow$ & odsF$\uparrow$\\
\midrule
\multicolumn{5}{c}{\textit{CNN-based decoder}} \\
Cross-Stitch \cite{misra2016cross} & 36.34 & 0.6290 & 20.88 & 76.38\\
PAP \cite{zhang2019pattern} & 36.72 & 0.6178 & 20.82 & 76.42\\
PSD \cite{zhou2020pattern} & 36.69 & 0.6246 & 20.87 & 76.42 \\
PAD-Net \cite{xu2018pad} & 36.61 & 0.6270 & 20.85 & 76.38 \\
MTI-Net \cite{vandenhende2020mti} & 45.97 & 0.5365 & 20.27 & 77.86 \\
ATRC \cite{bruggemann2021exploring} & 46.33 & 0.5363 & 20.18 & 77.94 \\
\midrule
\multicolumn{5}{c}{\textit{Transformer-based decoder}} \\
InvPT \cite{ye2022inverted} & 53.56 & {0.5183} & {19.04} & 78.10\\
InvPT++ \cite{ye2023invptplus} & 53.85 & 0.5096 & 18.67 & 78.10 \\
TaskPrompter \cite{ye2023taskprompter} & 55.30 & 0.5152 & {18.47} & 78.20 \\
MQTransformer \cite{xu2023multi} & {54.84} & 0.5325 & 19.67 & {78.20}\\
{TSP-Transformer \cite{wang2024tsp}} & {55.39} & {\underline{0.4961}} & {18.44} & {77.50}\\
{MLoRE \cite{yang2024multi}} & {\underline{55.96}} & {0.5076} & {\underline{18.33}} & {78.43}\\
\midrule
\multicolumn{5}{c}{\textit{Diffusion-based decoder}} \\
{TaskDiffusion \cite{yang2025multi}} & {55.65} & {0.5020} & {{18.43}} & {78.64}\\
\midrule
\multicolumn{5}{c}{\textit{Mamba-based decoder}} \\
MTMamba \cite{lin2024mtmamba} & {55.82} & {0.5066} & {18.63} & \underline{78.70}\\
MTMamba++ & \textbf{57.01} & \textbf{0.4818} & \textbf{18.27} & \textbf{79.40} \\
\bottomrule
\end{tabular}}
\hfill
\resizebox{0.53\linewidth}{!}{
\begin{tabular}{lccccc}
\toprule
\multirow{2}{*}{\textbf{Method}} & \textbf{Semseg} & \textbf{Parsing} & \textbf{Saliency} & \textbf{Normal} & \textbf{Boundary}\\
& mIoU$\uparrow$ & mIoU$\uparrow$ & maxF$\uparrow$ & mErr$\downarrow$ & odsF$\uparrow$\\
\midrule
\multicolumn{6}{c}{\textit{CNN-based decoder}} \\
ASTMT \cite{maninis2019attentive} & 68.00 & 61.10 & 65.70 & 14.70 & 72.40\\
PAD-Net \cite{xu2018pad} &  53.60 & 59.60 & 65.80 & 15.30 & 72.50 \\
MTI-Net \cite{vandenhende2020mti}  & 61.70 & 60.18 & 84.78 & 14.23 & 70.80 \\
ATRC \cite{bruggemann2021exploring}  & 62.69 & 59.42 & 84.70 & 14.20 & 70.96 \\
ATRC-ASPP \cite{bruggemann2021exploring} & 63.60 & 60.23 & 83.91 & 14.30 & 70.86 \\
ATRC-BMTAS \cite{bruggemann2021exploring} & 67.67 & 62.93 & 82.29 & 14.24 & 72.42 \\
\midrule
\multicolumn{6}{c}{\textit{Transformer-based decoder}} \\
InvPT \cite{ye2022inverted} & {79.03} & {67.61} & {84.81} & {14.15} & 73.00\\
InvPT++ \cite{ye2023invptplus} & 80.22 & 69.12 & 84.74 & {13.73} & 74.20\\
TaskPrompter \cite{ye2023taskprompter} & 80.89 & 68.89 & {84.83} & {13.72} & 73.50\\
MQTransformer \cite{xu2023multi} & 78.93 & 67.41 & 83.58 & 14.21 & {73.90}\\
{TSP-Transformer \cite{wang2024tsp}} & {\underline{81.48}} & {70.64} & {84.86} & {13.69} & {74.80}\\
{MLoRE \cite{yang2024multi}} & {81.41} & {70.52} & {84.90} & {\textbf{13.51}} & {75.42} \\
\midrule
\multicolumn{6}{c}{\textit{Diffusion-based decoder}} \\
{TaskDiffusion \cite{yang2025multi}} & {81.21} & {69.62} & {\underline{84.94}} & {\underline{13.55}} & {74.89} \\
\midrule
\multicolumn{6}{c}{\textit{Mamba-based decoder}} \\
MTMamba \cite{lin2024mtmamba} & {81.11} & \underline{72.62} & {84.14} & {14.14} & \textbf{78.80}\\
MTMamba++ & \textbf{81.94} & \textbf{72.87} & \textbf{85.56} & 14.29 & \underline{78.60}\\
\bottomrule
\end{tabular}}
\label{tab:pascal}
\end{table*}

\begin{table}[!t]
\centering
\caption{Comparison with state-of-the-art methods on the Cityscapes dataset. $\uparrow (\downarrow)$ indicates that a higher (lower) result corresponds to better performance.  The best and second-best results are highlighted in \textbf{bold} and \underline{underline}, respectively.}
\vspace{-0.1in}
\begin{tabular}{lcc}
\toprule
\multirow{2}{*}{\textbf{Method}} & \textbf{Semseg} & \textbf{Disparity}\\
& mIoU$\uparrow$ & RMSE$\downarrow$ \\
\midrule
\multicolumn{3}{c}{\textit{CNN-based decoder}} \\
PAD-Net \cite{xu2018pad} & 53.19 & 5.05\\
MTI-Net \cite{vandenhende2020mti} & 59.85 & 5.06\\
\midrule
\multicolumn{3}{c}{\textit{Transformer-based decoder}} \\
InvPT \cite{ye2022inverted} & 71.78 & 4.67 \\
TaskPrompter \cite{ye2023taskprompter} & 72.41 & 5.49\\
\midrule
\multicolumn{3}{c}{\textit{Mamba-based decoder}} \\
MTMamba \cite{lin2024mtmamba} & \underline{78.00} & \underline{4.66}\\
MTMamba++ & \textbf{79.13} & \textbf{4.63}\\
\bottomrule
\end{tabular}
\label{tab:cityscapes}
\end{table}

\subsubsection{Evaluation Metrics} 
Following \cite{ye2022inverted, ye2023invptplus, ye2023taskprompter}, we adopt mean intersection over union (mIoU) as the evaluation metric for semantic segmentation and human parsing tasks, root mean square error (RMSE) for monocular depth estimation and disparity estimation tasks, mean error (mErr) for surface normal estimation task, maximal F-measure (maxF) for saliency detection task, and optimal-dataset-scale F-measure (odsF) for object boundary detection task. Moreover, we report the average relative performance improvement of an MTL model $\hA$ over single-task (STL) models as the overall metric, which is defined as follows,
\begin{equation}
\Delta_m(\hA) = 100\%\times \frac{1}{T}\sum_{t=1}^{T}(-1)^{s_t}\frac{M_{t}^\hA-M_{t}^{\text{STL}}}{M_{t}^{\text{STL}}},
\end{equation}
where $T$ is the number of tasks, $M_{t}^\hA$ is the metric value of method $\hA$ on task $t$, and $s_{t}$ is $0$ if a larger value indicates better performance for task $t$, and $1$ otherwise.

\subsection{Comparison with State-of-the-art Methods}

We compare the proposed MTMamba++ method with two types of MTL methods: 
\begin{enumerate*}[(i), series = tobecont, itemjoin = \quad]
\item CNN-based methods, including Cross-Stitch \cite{misra2016cross}, PAP \cite{zhang2019pattern}, PSD \cite{zhou2020pattern}, PAD-Net \cite{xu2018pad}, MTI-Net \cite{vandenhende2020mti}, ATRC \cite{bruggemann2021exploring}, and ASTMT \cite{maninis2019attentive};
\item Transformer-based methods, including InvPT \cite{ye2022inverted}, TaskPrompter \cite{ye2023taskprompter},  InvPT++ \cite{ye2023invptplus}, MQTransformer \cite{xu2023multi}, {TSP-Transformer \cite{wang2024tsp}, and MLoRE \cite{yang2024multi}}; and  
\item {Diffusion-based method TaskDiffusion \cite{yang2025multi}}. 
\end{enumerate*}

Table \ref{tab:pascal} provides the results on NYUDv2 and PASCAL-Context datasets. {As can be seen, MTMamba++ largely outperforms CNN-based, Transformer-based, and Diffusion-based methods, especially achieving the best performance in all four tasks of NYUDv2. Notably, MTMamba++ shows significant improvements over MLoRE \cite{yang2024multi} by +1.05 (mIoU) and +0.97 (odsF) in semantic segmentation and object boundary detection tasks, which demonstrates the superiority of MTMamba++.}  Moreover, MTMamba++ performs better than MTMamba, showing the effectiveness of S-CTM and LiteHead.

{On the PASCAL-Context dataset, MTMamba++ continues to demonstrate superior performance on all tasks except the normal prediction task, which is also comparable. Compared with MLoRE \cite{yang2024multi}, MTMamba++ achieves notable improvements of +0.53 (mIoU), +2.35 (mIoU), +0.66 (maxF), and +3.18 (odsF) in semantic segmentation, human parsing, saliency detection, and object boundary detection tasks, respectively. When compared to the diffusion-based method TaskDiffusion \cite{yang2025multi}, MTMamba++ shows advantages of +0.73 (mIoU), +3.25 (mIoU), +0.62 (maxF), and +3.71 (odsF) in four tasks. These results clearly demonstrate the effectiveness of MTMamba++ for multi-task dense prediction.} Furthermore, MTMamba++ outperforms our preliminary work MTMamba on three of five tasks while maintaining comparable performance on the remaining two, further validating the effectiveness of our proposed components.

Table \ref{tab:cityscapes} shows the results on the Cityscapes dataset. We can see that Mamba-based methods perform largely better than the previous CNN-based and Transformer-based approaches on both two tasks. Moreover, MTMamba++ archives the best performance.
Notably, MTMamba++ outperforms TaksPrompter \cite{ye2023taskprompter} by +6.72 (mIoU) in the semantic segmentation task, demonstrating that MTMamba++ is more effective. 
Besides, MTMamba++ performs better than MTMamba, which shows the effectiveness of S-CTM and LiteHead.

\begin{table*}[!t]
\centering
\caption{Effectiveness of each core component on NYUDv2. ``Multi-task'' denotes an MTL model where each task uses standard Swin Transformer blocks \cite{liu2021swin} after the ECR block in each decoder stage. ``Single-task'' is the single-task counterpart of ``Multi-task''. {\#11} is the default configuration of MTMamba++. }
\vspace{-0.1in}
\begin{tabular}{cllc|ccccc|cc}
\toprule
\multirow{2}{*}{\#} & 
\multirow{2}{*}{\textbf{Method}} & \multirow{2}{*}{\textbf{Each Decoder Stage}} & \multirow{2}{*}{\textbf{Head}}
& \textbf{Semseg} & \textbf{Depth} & \textbf{Normal} & \textbf{Boundary} & \bm{$\Delta_m$}[\%] & \textbf{\#Param} & \textbf{FLOPs}\\
&& & & mIoU$\uparrow$ & RMSE$\downarrow$ & mErr$\downarrow$ & odsF$\uparrow$ & $\uparrow$ & MB$\downarrow$ & GB$\downarrow$\\
\midrule
1 & \multirow{2}{*}{Single-task} & 2*Swin & DenseHead & 54.32 & 0.5166 & 19.21 & 77.30 & 0.00 & 889 & 1075\\
{2} &  & {2*STM} & {DenseHead} & {54.94} &{0.5100} & {18.85} & {78.00} & {+1.29} & {864} & {1040}\\
\midrule
3 & \multirow{4}{*}{Multi-task} & 2*Swin & DenseHead & 53.72 & 0.5239 & 19.97 & 76.50 & -1.87 & 303 & 466\\
{4} & & {2*Swin} & {LiteHead} & {53.37} & {0.5201} & {19.62} & {78.40} & {-0.78} & {302} & {436}\\
{5} & & {3*Swin} & {DenseHead} & {54.22} & {0.5225} & {19.84} & {77.40} & {-1.11} & {341} & {563}\\
{6} & & {3*Swin} & {LiteHead} & {54.44} & {0.5117} & {19.65} & {78.60} & {+0.14} & {339} & {533}\\
\midrule
7 & \multirow{5}{*}{MTMamba++} & 2*STM & DenseHead &  54.66 & {0.4984} & 18.81 & 78.20 & +1.84 & 276 & 435\\
8 & & 3*STM & DenseHead & 54.75 & 0.5054 & 18.81 & 78.20 & +1.55 & 300 & 517\\
9 & & 2*STM+1*F-CTM & DenseHead & {55.82} & {0.5066} & {18.63} & {78.70} & {+2.38} & 308 & 541\\
10 & & 2*STM+1*F-CTM & LiteHead & 56.53 & 0.5054 & 18.71 & 79.20 & +2.82 & 306 & 510\\
11 & & 2*STM+1*S-CTM & LiteHead & \textbf{57.01} & \textbf{0.4818} & \textbf{18.27} & \textbf{79.40} & \textbf{+4.82} & 315 & 524 \\
\bottomrule
\end{tabular}
\label{tab:block_ablation}
\end{table*}

The qualitative comparisons with baselines (i.e., InvPT \cite{ye2022inverted}, TaskPrompter \cite{ye2023taskprompter}, and MTMamba \cite{lin2024mtmamba}) on NYUDv2, PASCAL-Context, and Cityscapes datasets are shown in Figures \ref{fig:qualitative_nyu}, \ref{fig:qualitative_pascal}, and \ref{fig:qualitative_city}, demonstrating that MTMmaba++ provides more precise predictions and details.

\subsection{Model Analysis}

{In this section, we provide a comprehensive analysis of the proposed MTMamba++. Without specific instructions, the encoder in this section is the Swin-Large Transformer.}

\subsubsection{Effectiveness of Each Component}
The decoders of MTMamba++ contain two types of core blocks: STM and CTM blocks. 
Compared to the preliminary version MTMamba \cite{lin2024mtmamba}, MTMamba++ replaces the F-CTM block and DenseHead of MTMamba with the S-CTM block and LiteHead, respectively. 

In this experiment, we study the effectiveness of each component on the NYUDv2 dataset. {We first introduce two groups of baselines:
\begin{enumerate*}[(i), series = tobecont, itemjoin = \quad] 
\item ``Multi-task" represents an MTL model using only standard Swin Transformer blocks \cite{liu2021swin} after the ECR block in each decoder stage for each task; and 
\item ``Single-task'' means that each task has a task-specific encoder-decoder. 
\end{enumerate*}
}
The results are shown in Table \ref{tab:block_ablation}, where {\#9} and {\#11} are the default configurations of MTMamba and MTMamba++, respectively. 

Firstly, the STM block outperforms the Swin Transformer block \cite{liu2021swin} in terms of efficiency and effectiveness for multi-task dense prediction, as indicated by the superior results in Table \ref{tab:block_ablation} ({\#3} vs. {\#7 and \#5 vs. \#8}). 
Secondly, merely increasing the number of STM blocks from two to three does not enhance performance significantly. When the F-CTM block is incorporated, the performance largely improves in terms of $\Delta_m$ ({\#9} vs. {\#7}/{\#8}), demonstrating the effectiveness of F-CTM. 
Thirdly, {comparisons between \#3 and \#4, \#5 and \#6, as well as \#9 and \#10 show} that LiteHead is more effective and efficient than DenseHead.
Fourthly, compared {\#10} with {\#11}, we can find that replacing F-CTM with S-CTM leads to a significant performance improvement in all tasks with a tiny additional cost,
demonstrating that the semantic-aware interaction in S-CTM is more effective than F-CTM.  
Finally, the default configuration of MTMamba++ significantly surpasses the ``Single-task'' baselines across all tasks ({\#11 vs. \#1/\#2}), thereby demonstrating the effectiveness of MTMamba++.

\subsubsection{Comparison between SSM and Attention}

To demonstrate the superiority of the SSM-based architecture in multi-task dense prediction, we conduct an experiment on NYUDv2 by replacing the SSM-related components in MTMamba++ with attention-based counterparts. Specifically, we substitute the SS2D module in the STM block with window-based multi-head self-attention \cite{liu2021swin} and replace the CSS2D module in the S-CTM block with window-based multi-head cross-attention. {The comparative results in Table \ref{tab:attention_ablation} show that MTMamba++ significantly outperforms the attention-based variant across all tasks while requiring approximately 29.7\% fewer parameters and 34.2\% lower FLOPs. This efficiency advantage is primarily from SSM's linear computational complexity with respect to sequence length, in contrast to the quadratic complexity of attention mechanisms. These results demonstrate that SSM-based architectures are more effective and efficient for multi-task dense prediction tasks, where we need to process high-resolution feature maps in pixel-level prediction.}

\subsubsection{Effectiveness of Each Decoder Stage} 
\begin{figure}[!t]
\centering
\includegraphics[width=\linewidth]{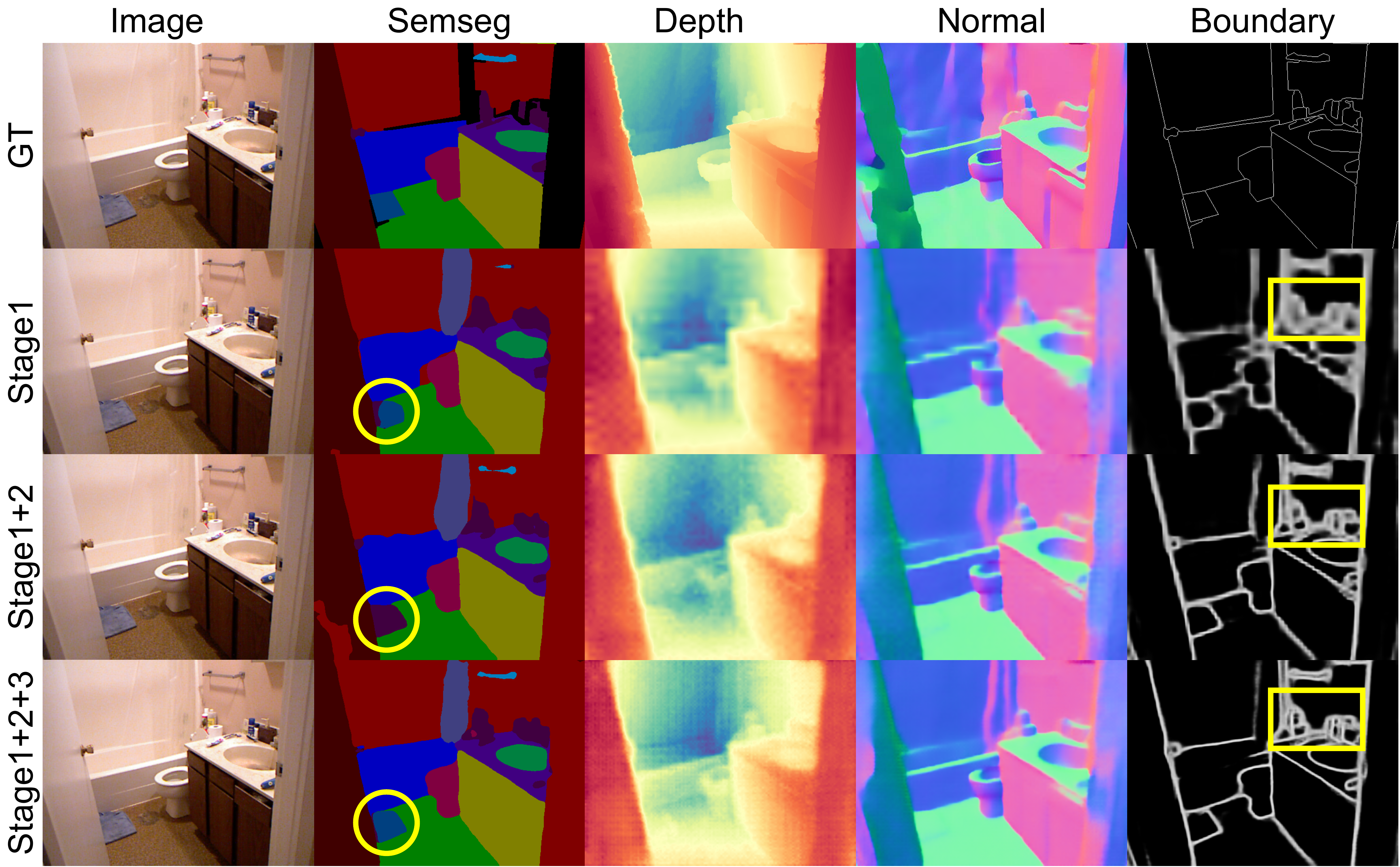}
\vspace{-0.25in}
\caption{{A qualitative comparison of each decoder stage in MTMamba++ on NYUDv2. Zoom in for more details.}}
\label{fig:nyu_stage}
\end{figure}

As shown in Figure \ref{fig:overall_arch}, 
the decoder of MTMamba++ consists of three stages. 
In this experiment, we study the effectiveness of these three stages on the NYUDv2 dataset. 
Table \ref{tab:stage_ablation} presents the ablation results, clearly demonstrating that each decoder stage contributes positively to the performance of MTMamba++. {The progressive performance gains achieved by successively incorporating each stage validate the effectiveness of our multi-stage decoder design in capturing and integrating multi-scale contextual features. As visualized in Figure \ref{fig:nyu_stage}, this hierarchical feature aggregation enables progressively refined predictions with sharper boundaries, particularly benefiting the boundary detection task.}

\begin{table}[!t]
\centering
\caption{Comparison between SSM and attention on NYUDv2. We replace the SSM-related modules in MTMamba++ (i.e., the SS2D and CSS2D modules) with attention-based mechanisms (i.e., self-attention and cross-attention mechanisms). 
}
\tabcolsep=0.1cm
\vspace{-0.1in}
\resizebox{\linewidth}{!}{
\begin{tabular}{c|ccccc|cc}
\toprule 
& \textbf{Semseg} & \textbf{Depth} & \textbf{Normal} & \textbf{Boundary} & \multirow{1}{*}{\bm{$\Delta_m$}[\%]} & \textbf{\#Param} & \textbf{FLOPs}\\
& mIoU$\uparrow$ & RMSE$\downarrow$ & mErr$\downarrow$ & odsF$\uparrow$ & $\uparrow$ & MB$\downarrow$ & GB$\downarrow$\\
\midrule
attention-based & 55.15 & 0.4945 & 18.72 & 79.00 & +2.63 & 448 & 796\\
SSM-based & \textbf{57.01} & \textbf{0.4818} & \textbf{18.27} & \textbf{79.40} & \textbf{+4.82} & 315 & 524\\
\bottomrule
\end{tabular}}
\label{tab:attention_ablation}
\end{table}

\begin{table}[!t]
\centering
\tabcolsep=0.1cm
\caption{Effectiveness of each decoder stage in MTMamba++ on NYUDv2. }
\vspace{-0.1in}
\resizebox{\linewidth}{!}{
\begin{tabular}{ccc|ccccc|cc}
\toprule
\multirow{2}{*}{\textbf{Stage1}} & \multirow{2}{*}{\textbf{Stage2}}  & \multirow{2}{*}{\textbf{Stage3}} 
& \textbf{Semseg} & \textbf{Depth} & \textbf{Normal} & \textbf{Boundary} & \multirow{1}{*}{\bm{$\Delta_m$}[\%]} & \textbf{\#Param} & \textbf{FLOPs}\\
& & & mIoU$\uparrow$ & RMSE$\downarrow$ & mErr$\downarrow$ & odsF$\uparrow$ & $\uparrow$ & MB$\downarrow$ & GB$\downarrow$\\
\midrule
\Checkmark & \XSolidBrush & \XSolidBrush & 55.50 & 0.4960 & 18.98 & 67.90 & -1.20 & 287 & 291\\
\Checkmark & \Checkmark & \XSolidBrush & 55.54 & 0.4872 & 18.46 & 77.70 & +3.08 & 309 & 393\\
\Checkmark & \Checkmark & \Checkmark & \textbf{57.01} & \textbf{0.4818} & \textbf{18.27} & \textbf{79.40} & \textbf{+4.82} & 315 & 524\\
\bottomrule
\end{tabular}}
\label{tab:stage_ablation}
\end{table}

\begin{table}[!t]
\centering
\caption{Effect of each scan in CSS2D module on NYUDv2.}
\vspace{-0.1in}
\resizebox{\linewidth}{!}{
\begin{tabular}{c|ccccc}
\toprule
& \textbf{Semseg} & \textbf{Depth} & \textbf{Normal} & \textbf{Boundary} & \multirow{1}{*}{\bm{$\Delta_m$}[\%]} \\
& mIoU$\uparrow$ & RMSE$\downarrow$ & mErr$\downarrow$ & odsF$\uparrow$ & $\uparrow$\\
\midrule
MTMamba++ & \textbf{57.01} & \textbf{0.4818} & {18.27} & \textbf{79.40} & \textbf{+4.82}\\
\quad w/o scan1 & 56.02 & 0.4962 & 18.30 & 79.30 & +3.60 \\
\quad w/o scan2 & 56.50 & 0.4967 & \textbf{18.22} & 79.30 & +3.90\\
\quad w/o scan3 & 56.09 & 0.4874 & 18.41 & 79.30 & +3.91\\
\quad w/o scan4 & 56.27 & 0.4942 & 18.36 & 79.30 & +3.73\\
\bottomrule
\end{tabular}}
\label{tab:direction_ablation}
\end{table}

\begin{table}[!h]
\centering
\caption{Effect of expand factor $\alpha$ in MTMamba++ on NYUDv2 {with different numbers of tasks. ``S'', ``D'', ``N'', and ``B'' denote the semantic segmentation, depth estimation, surface normal estimation, and boundary detection tasks, respectively.}}
\vspace{-0.1in}
\resizebox{\linewidth}{!}{
\begin{tabular}{c|c|ccccc}
\toprule 
& \multirow{2}{*}{$\alpha$} & \textbf{Semseg} & \textbf{Depth} & \textbf{Normal} & \textbf{Boundary} & \multirow{1}{*}{\bm{$\Delta_m$}[\%]} \\
& & mIoU$\uparrow$ & RMSE$\downarrow$ & mErr$\downarrow$ & odsF$\uparrow$ & $\uparrow$ \\
\midrule
\multirow{3}{*}{{S-D}} & {$1$} & {58.10} & {\textbf{0.4768}} & - & - & {\textbf{+7.33}}\\
& {$2$} & {\textbf{58.25}} & {{0.4808}} & - & - & {{+7.08}}\\
& {$3$} & {58.23} & {0.4859} & - & - & {+6.57}\\
\midrule
\multirow{3}{*}{{S-D-N}} & {$1$} & {54.85} & {0.4956} & {18.57} & - & {+2.79}\\
& {$2$} & {\textbf{55.78}} & {\textbf{0.4888}} & {18.43} & - & {\textbf{+4.04}}\\
& {$3$} & {55.30} & {0.4932} & {\textbf{18.39}} & - & {+3.53}\\
\midrule
\multirow{3}{*}{S-D-N-B} & $1$ & 56.74 & 0.4927 & 18.45 & \textbf{79.40} & +3.93 \\
& $2$ & \textbf{57.01} & \textbf{0.4818} & \textbf{18.27} & \textbf{79.40} & \textbf{+4.82} \\
& $3$ & 55.85 & 0.4882 & 18.40 & 79.10 & +3.71 \\
\bottomrule
\end{tabular}}
\label{tab:alpha_ablation}
\end{table}

\subsubsection{Effect of Each Scan in CSS2D Module}

As mentioned in Section \ref{sec:S-CTM}, the CSS2D module scans the 2D feature map from four different directions. We conduct an experiment on NYUDv2 to study the effect of each scan. The results are presented in Table \ref{tab:direction_ablation}. As can be seen, dropping any direction leads to a performance drop compared with the default configuration that uses all directions, showing that all directions are beneficial to MTMamba++. 

\subsubsection{Analysis of $\alpha$}

As mentioned in Sections \ref{sec:stm} and \ref{sec:S-CTM}, in MTMamba++, both STM and S-CTM blocks expand the feature channel to improve the model capacity by a hyperparameter $\alpha$. 
We conduct an experiment on NYUDv2 to {explore the relationship between $\alpha$ and task conflicts}. 
{Increasing the expansion factor $\alpha$ enhances the model's representational capacity for capturing both task-specific features and cross-task interactions. However, excessively large values can lead to increased computational complexity and over-parameterization. The redundancy in the representation space dilutes effective information and makes model optimization more challenging, resulting in worse performance.}

{
The results in Table \ref{tab:alpha_ablation} demonstrate that the optimal $\alpha$ value is correlated with the severity of task conflicts. For the 2-task setting (S-D), $\alpha=1$ achieves the best $\Delta_m$, because the semantic segmentation and depth estimation tasks have relatively low conflict, requiring minimal additional capacity for cross-task interaction modeling. However, when the normal estimation task is added in the 3-task setting (S-D-N), task conflicts become more severe as evidenced by the significant performance drop of both semantic segmentation and depth estimation tasks. In this case, $\alpha=2$ becomes optimal in terms of $\Delta_m$, indicating that increased representational capacity is needed to handle the heightened task conflicts. In the 4-task setting (S-D-N-B), while the boundary detection task is added, the conflicts appear to be somewhat alleviated as the boundary detection task can provide complementary information to other tasks. Thus, $\alpha=2$ continues to perform best in terms of $\Delta_m$, maintaining the balance between adequate capacity for conflict resolution and avoiding over-parameterization. Notably, $\alpha=3$ consistently underperforms across all configurations, demonstrating that excessively large expansion factors lead to over-parameterization.} 

{
These results demonstrate that smaller $\alpha$ suffices for low-conflict scenarios, while moderately larger $\alpha$ is beneficial when severe conflicts exist, but excessively large $\alpha$ always degrades performance. Thus, $\alpha=2$ is adopted as the default configuration in MTMamba++ as it provides robust performance across various multi-task scenarios.}

\begin{figure*}[!t]
\centering
\includegraphics[width=.49\linewidth]{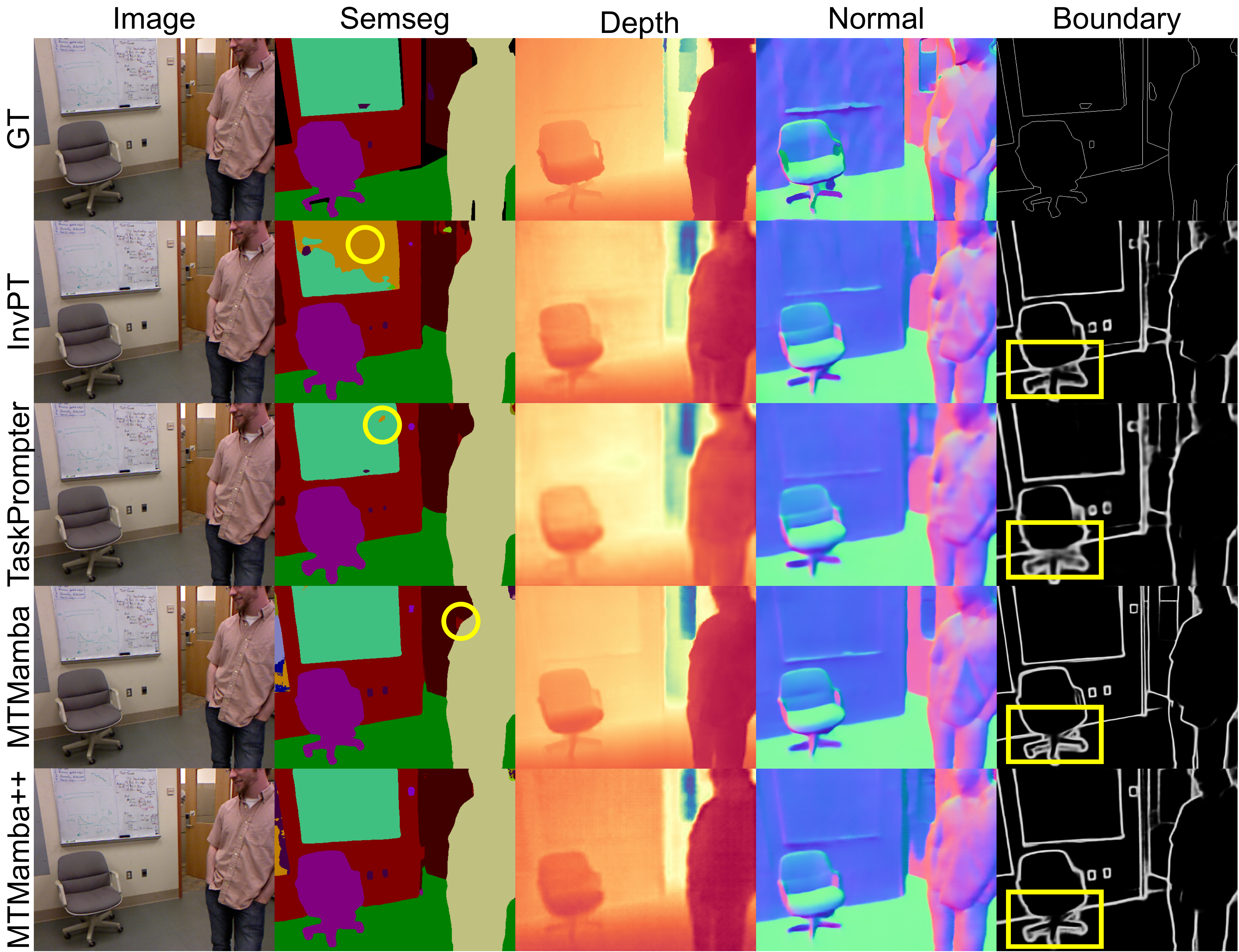}
\includegraphics[width=.49\linewidth]{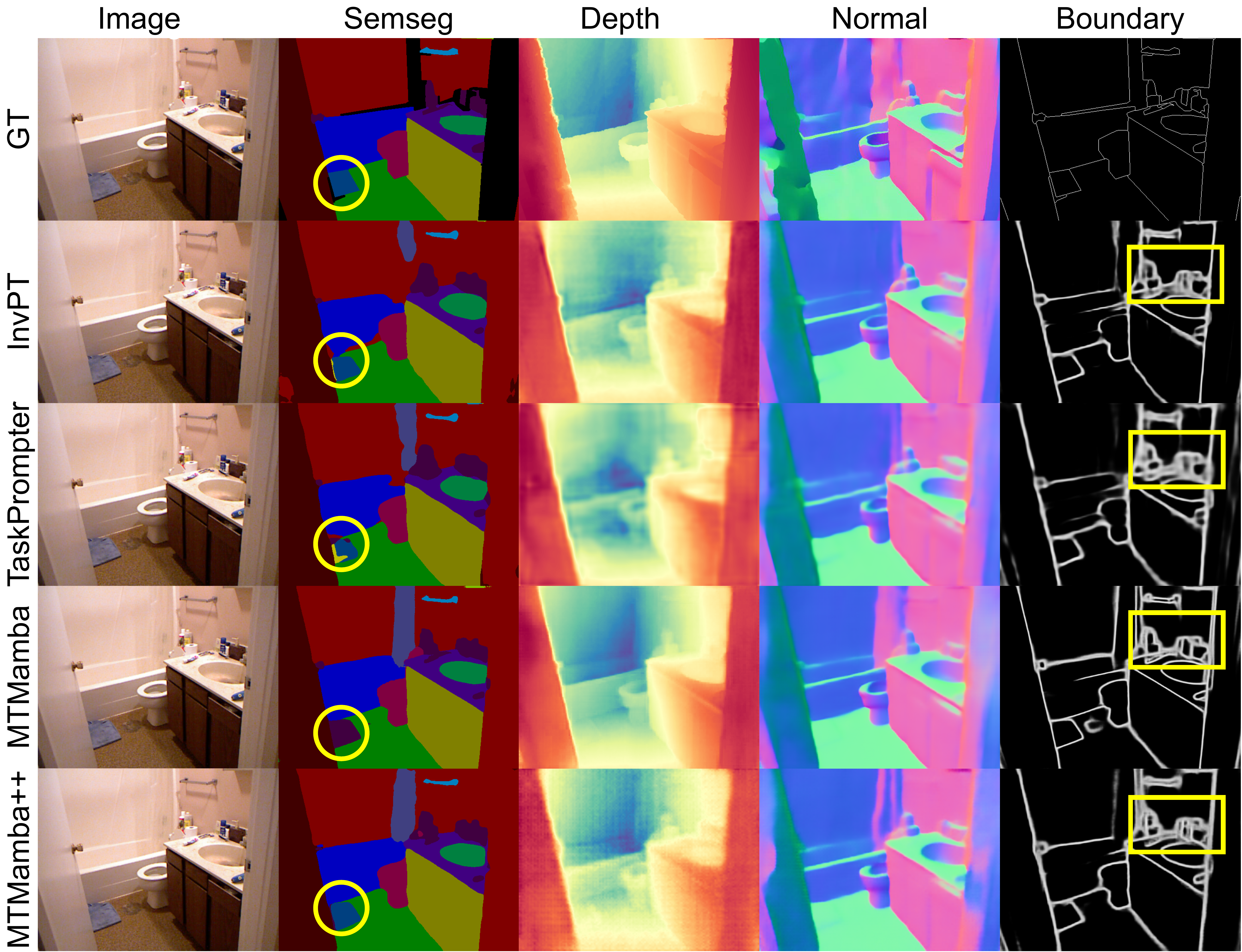}
\vspace{-0.1in}
\caption{Qualitative comparison with baselines (i.e., InvPT \cite{ye2022inverted}, TaskPrompter \cite{ye2023taskprompter}, and MTMamba \cite{lin2024mtmamba}) on the NYUDv2 dataset. As highlighted, MTMamba++ generates better predictions with more accurate details {and sharper boundaries}. In the semantic segmentation task, the black regions in GT denote the background and are excluded from the computation of loss and evaluation metric (i.e., mIoU). Zoom in for more details.}
\label{fig:qualitative_nyu}
\end{figure*}

\begin{table}[!t]
	\centering
	\tabcolsep=0.13cm
	\caption{Performance of MTMamba++ with different scales of Swin Transformer encoder on NYUDv2. }
	\vspace{-0.1in}
	\resizebox{\linewidth}{!}{
		\begin{tabular}{lccccc}
			\toprule
			\multirow{2}{*}{\textbf{Method}} & \multirow{2}{*}{\textbf{Encoder}}
			& \textbf{Semseg} & \textbf{Depth} & \textbf{Normal} & \textbf{Boundary}\\
			& & mIoU$\uparrow$ & RMSE$\downarrow$ & mErr$\downarrow$ & odsF$\uparrow$ \\
			\midrule
			MTMamba \cite{lin2024mtmamba} & \multirow{2}{*}{Swin-Small} & 51.93 & 0.5246 & \textbf{19.45} & 77.80\\
			MTMamba++ & & \textbf{52.44} & \textbf{0.5210} & 19.51 & \textbf{78.10}\\
			\midrule
			MTMamba \cite{lin2024mtmamba} & \multirow{2}{*}{Swin-Base} & 53.62 & 0.5126 & 19.28 & 77.70\\
			MTMamba++ & & \textbf{55.08} & \textbf{0.5006} & \textbf{18.78} & \textbf{78.60}\\
			\midrule
			MTMamba \cite{lin2024mtmamba} & \multirow{2}{*}{Swin-Large} & 55.82 & 0.5066 & 18.63 & 78.70\\
			MTMamba++ & & \textbf{57.01} & \textbf{0.4818} & \textbf{18.27} & \textbf{79.40}\\
			\bottomrule
	\end{tabular}}
	\label{tab:backbone}
\end{table}

\begin{table}[!t]
	\centering
	\tabcolsep=0.08cm
	\caption{Comparison with state-of-the-art methods in model size and cost on the PASCAL-Context dataset. $\dag$ denotes that the results are from \cite{ye2023taskprompter}.}
	\vspace{-0.1in}
	\resizebox{\linewidth}{!}{
		\begin{tabular}{l|cc|ccccc}
			\toprule
			\multirow{2}{*}{\textbf{Method}} & \textbf{\#Param} & \textbf{FLOPs} & \textbf{Semseg} & \textbf{Parsing} & \textbf{Saliency} & \textbf{Normal} & \textbf{Boundary}\\
			& MB$\downarrow$ & GB$\downarrow$ & mIoU$\uparrow$ & mIoU$\uparrow$ & maxF$\uparrow$ & mErr$\downarrow$ & odsF$\uparrow$ \\
			\midrule
			PAD-Net$^\dag$ \cite{xu2018pad} & 330 & 773 & 78.01 & 67.12 & 79.21 & 14.37 & 72.60\\
			MTI-Net$^\dag$ \cite{vandenhende2020mti} & 851 & 774 & 78.31 & 67.40 & 84.75 & 14.67 & 73.00\\
			ATRC$^\dag$ \cite{bruggemann2021exploring} & 340 & 871 & 77.11 & 66.84 & 81.20 & 14.23 & 72.10\\
			InvPT$^\dag$ \cite{ye2022inverted} & 423 & 669 & {79.03} & {67.61} & {84.81} & {14.15} & 73.00\\
			TaskPrompter$^\dag$ \cite{ye2023taskprompter} & 401 & {497} & 80.89 & 68.89 & 84.83 & {13.72} & 73.50\\
			InvPT++ \cite{ye2023invptplus} & 421 & 667 & 80.22 & 69.12 & 84.74 & 13.73 & 74.20\\
			{TSP-Transformer \cite{wang2024tsp}} & {422} & {1991} & {81.48} & {70.64} & {84.86} & {13.69} & {74.80}\\
			{MLoRE \cite{yang2024multi}} & {407} & {571} & {81.41} & {70.52} & {84.90} & {\textbf{13.51}} & {75.42} \\
			{TaskDiffusion \cite{yang2025multi}} & {416} & {610} & {81.21} & {69.62} & {84.94} & {13.55} & {74.89} \\
			\midrule
			MTMamba \cite{lin2024mtmamba} & 336 & 632 & 81.11 & 72.62 & 84.14 & 14.14 & \textbf{78.80}\\
			MTMamba++ & 343 & {609}  & \textbf{81.94} & \textbf{72.87} & \textbf{85.56} & 14.29 & 78.60\\
			\bottomrule
	\end{tabular}}
	\label{tab:cost}
\end{table}

\subsubsection{Performance on Different Encoders} 
We perform an experiment on NYUDv2 to investigate the performance of the proposed MTMamba++ with different scales of Swin Transformer encoder. 
The results are shown in Table \ref{tab:backbone}. 
As can be seen, as the model capacity increases, MTMamba++ performs better on all tasks accordingly. 
Moreover, MTMamba++ consistently outperforms MTMamba on different encoders, confirming the effectiveness of the proposed S-CTM and LiteHead.

\subsubsection{Analysis of Model Size and Cost}
Table \ref{tab:cost} compares model size and FLOPs between the proposed MTMamba++ and baselines on the PASCAL-Context dataset. 
{We can see that MTMamba++ achieves state-of-the-art performance while maintaining high computational efficiency. Specifically, with only 343MB parameters (14.3\%, 18.7\%, 15.7\%, and 17.5\% fewer than InvPT, TSP-Transformer, MLoRE, and TaskDiffusion, respectively), our MTMamba++ achieves superior performance across most tasks. In terms of computational cost, MTMamba++ requires only 609GB FLOPs, which is merely 30.6\% of the resources needed by TSP-Transformer (1991GB) while still outperforming it. Compared to MLoRE and TaskDiffusion, MTMamba++ achieves better results with comparable computational demands. These results confirm that MTMamba++ offers not only performance advantages but also practical benefits for real-world applications through its efficient use of computational resources.}

\begin{figure*}[!t]
\centering
\includegraphics[width=.465\linewidth]{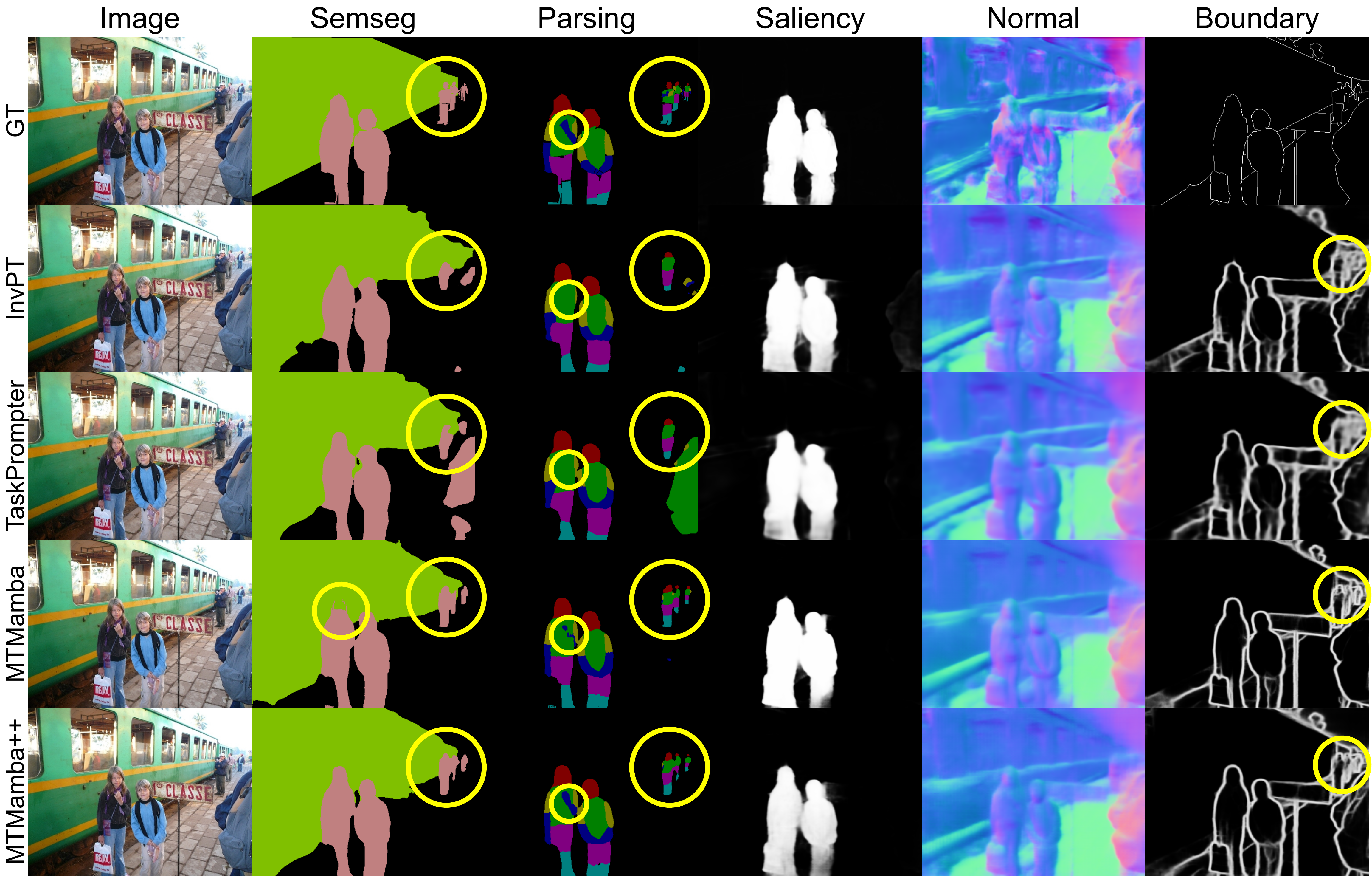}
\includegraphics[width=.523\linewidth]{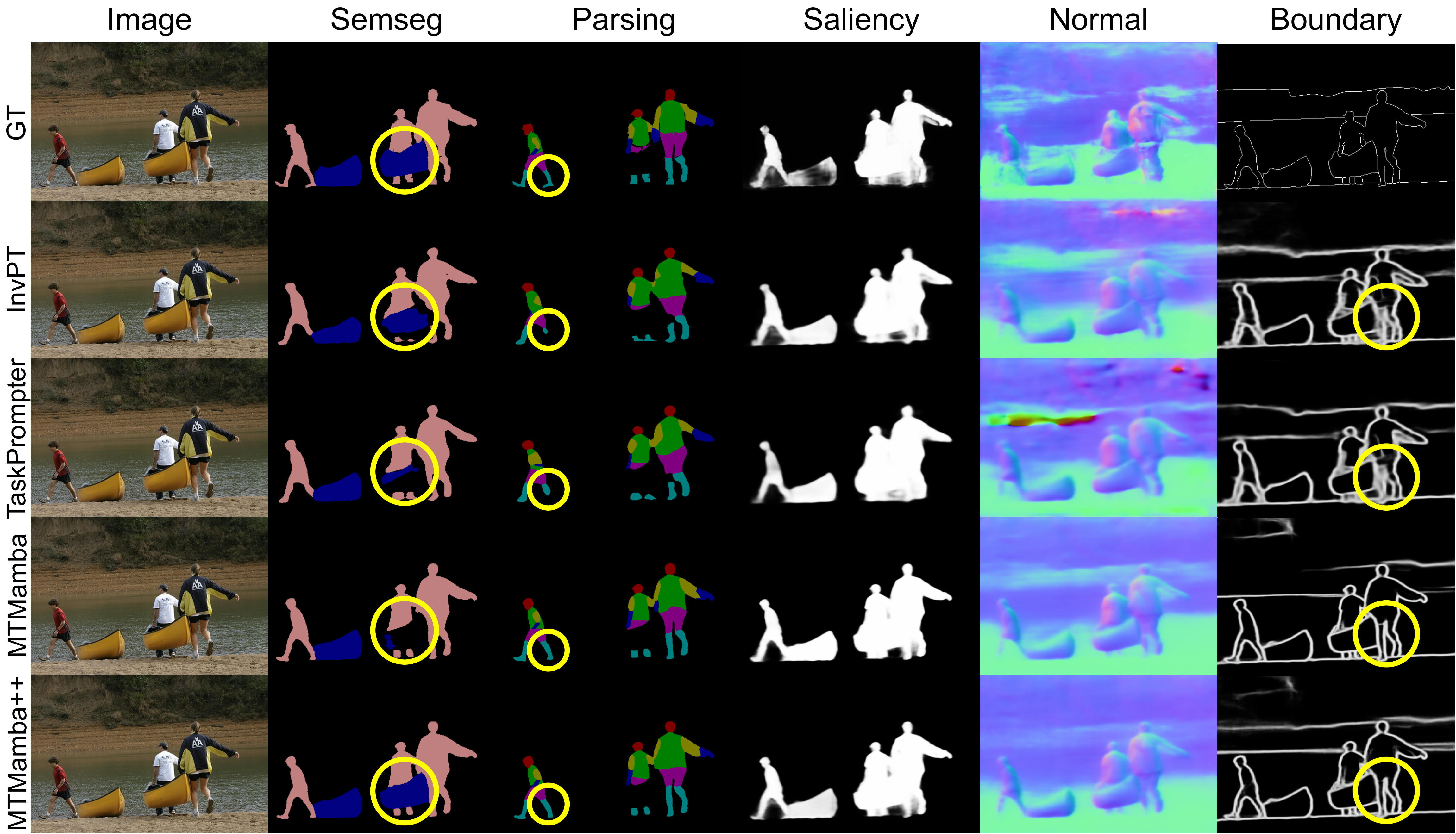}
\vspace{-0.1in}
\caption{Qualitative comparison with baselines (i.e., InvPT \cite{ye2022inverted}, TaskPrompter \cite{ye2023taskprompter}, and MTMamba \cite{lin2024mtmamba}) on the PASCAL-Context dataset. As highlighted, MTMamba++ generates better predictions with {sharper boundaries and greater precision in small objects}. Zoom in for more details.}
\label{fig:qualitative_pascal}
\end{figure*}

\begin{figure*}[!t]
\centering
\includegraphics[width=.48\linewidth]{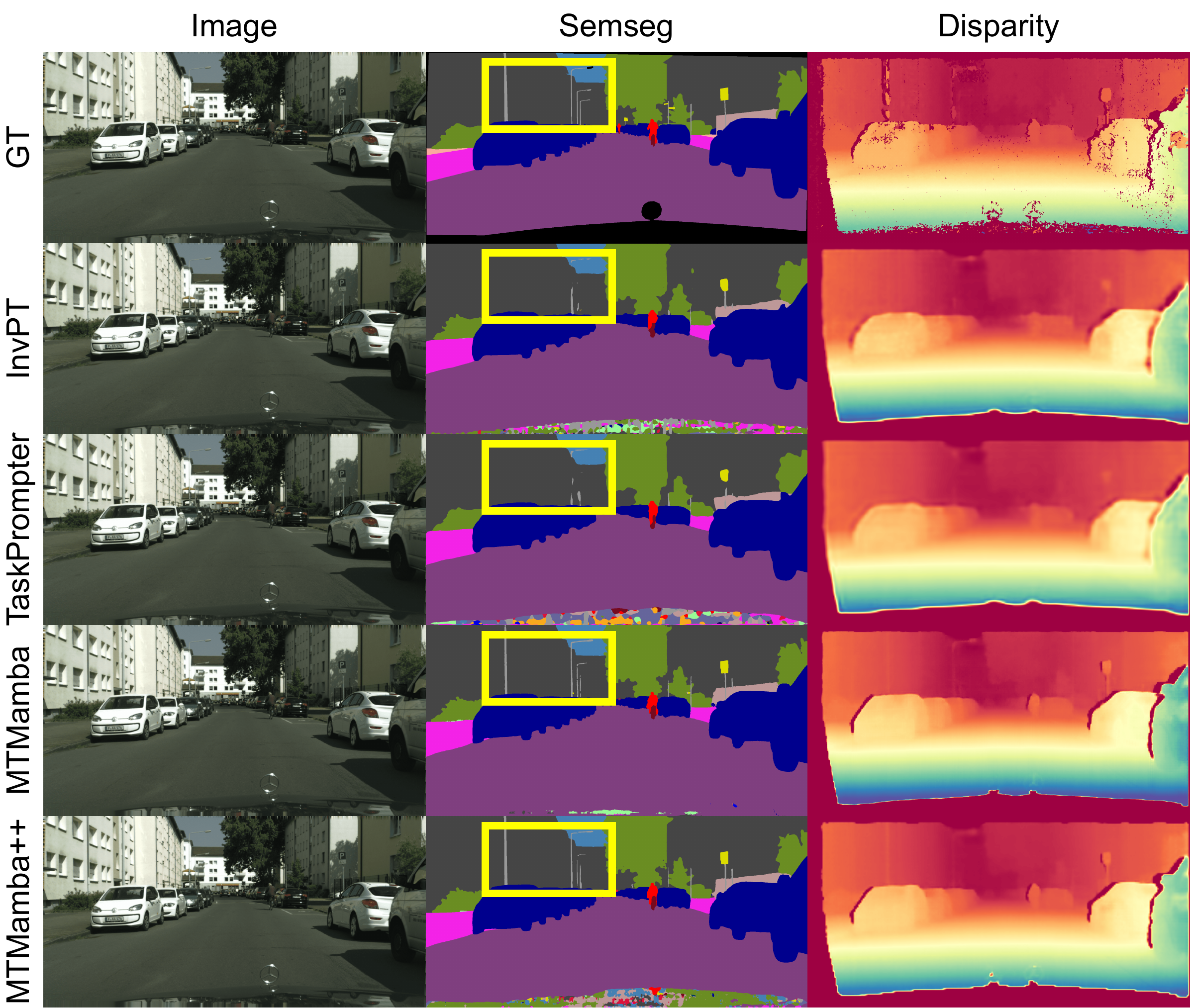}
\includegraphics[width=.48\linewidth]{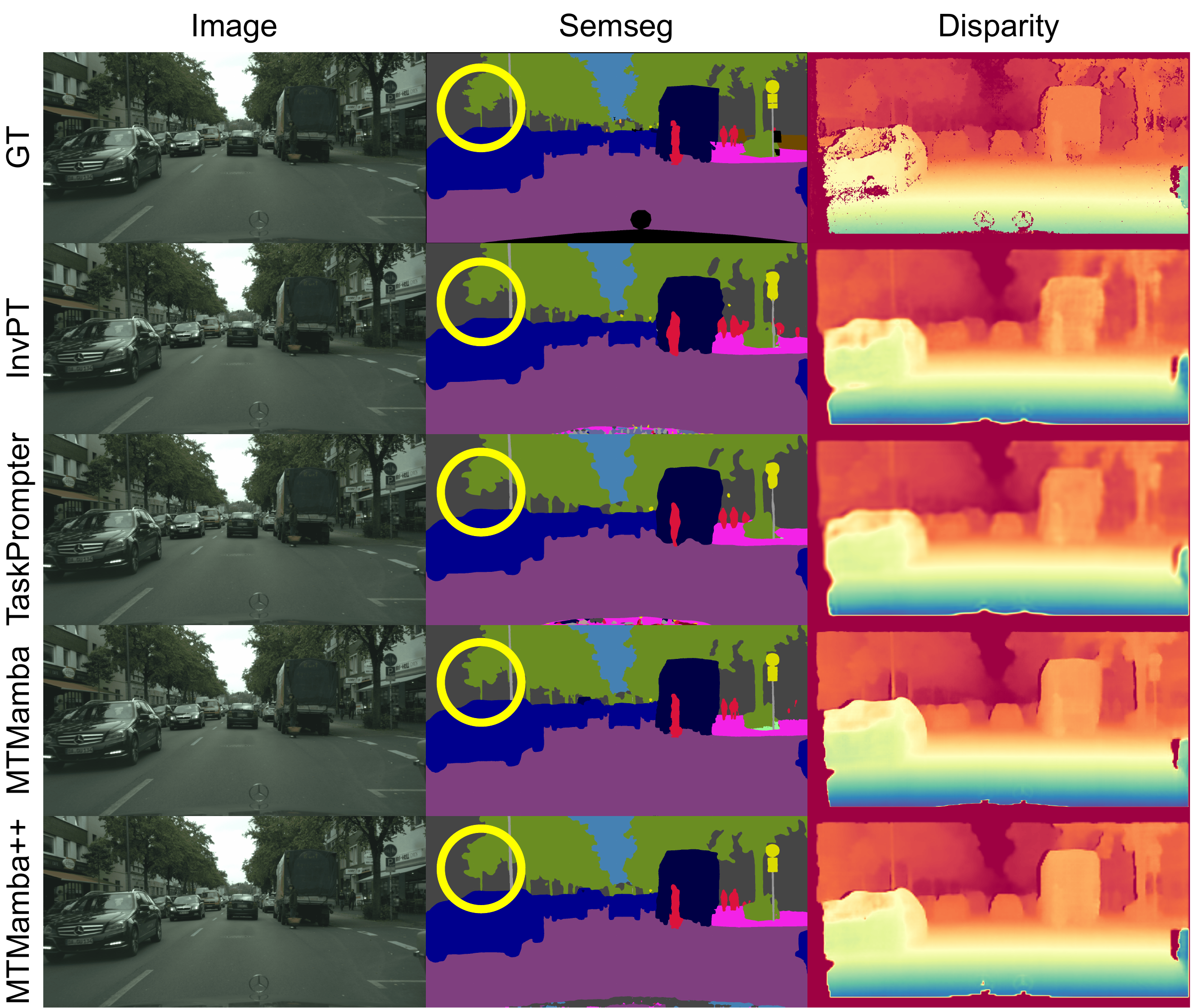}
\vspace{-0.1in} 
\caption{Qualitative comparison with baselines (i.e., InvPT \cite{ye2022inverted}, TaskPrompter \cite{ye2023taskprompter}, and MTMamba \cite{lin2024mtmamba}) on the Cityscapes dataset. As highlighted, MTMamba++ produces more precise predictions {in small objects}. Zoom in for more details.}
\label{fig:qualitative_city}
\end{figure*}

\subsection{Visualization of Predictions}

In this section, we compare the output predictions from MTMamba++ against baselines, including InvPT \cite{ye2022inverted}, TaskPrompter \cite{ye2023taskprompter}, and MTMamba \cite{lin2024mtmamba}. Figures \ref{fig:qualitative_nyu}, \ref{fig:qualitative_pascal}, and \ref{fig:qualitative_city} show the qualitative results on NYUDv2, PASCAL-Context, and Cityscapes datasets, respectively. As can be seen, MTMamba++ has better visual results than baselines in all datasets. 
{For example, as highlighted with yellow circles in Figure \ref{fig:qualitative_nyu}, MTMamba++ demonstrates fewer misclassification errors in semantic segmentation and produces sharper predicted boundaries in the boundary detection task. Figure \ref{fig:qualitative_pascal} illustrates that MTMamba++ achieves more accurate detection of small objects in both semantic segmentation and human parsing tasks, particularly evident in the highlighted regions where our method can effectively detect distant pedestrians. MTMamba++ also generates sharper predicted boundaries for the object boundary detection task. Similarly, as highlighted in Figure \ref{fig:qualitative_city}, 
MTMamba++ achieves higher precision in detecting small objects (e.g., street lamps and tree trunks) in semantic segmentation, which are missed by Transformer-based methods.} Hence, both qualitative study (Figures \ref{fig:qualitative_nyu}, \ref{fig:qualitative_pascal}, and \ref{fig:qualitative_city}) and quantitative study (Tables \ref{tab:pascal} and \ref{tab:cityscapes}) show the superior performance of MTMamba++.

\section{Conclusion} \label{sec:conclusion}
In this paper, we propose MTMamba++, a novel multi-task architecture with a Mamba-based decoder for multi-task dense scene understanding. With two types of core blocks (i.e., STM and CTM blocks), MTMamba++ can effectively model long-range dependency and achieve cross-task interaction.
We design two variants of the CTM block to promote knowledge exchange across tasks from the feature and semantic perspectives, respectively.
Experiments on three benchmark datasets demonstrate that MTMamba++ achieves better performance than previous methods {while maintaining high computational efficiency}.

\section*{Acknowledgments}
This work is supported by Guangzhou-HKUST(GZ) Joint Funding Scheme (No. 2024A03J0241) and Guangdong Provincial Key Lab of Integrated Communication, Sensing and Computation for Ubiquitous Internet of Things (No. 2023B1212010007).

\bibliography{reference}
\bibliographystyle{IEEEtran}

\end{document}

%% file: utils.tex
\newcommand{\hA}{\mathcal{A}}

\newcommand{\bR}{\mathbb{R}}

\newcommand{\vg}{{\bf g}}
\newcommand{\vh}{{\bf h}}

\newcommand{\vx}{{\bf x}}
\newcommand{\vy}{{\bf y}}
\newcommand{\vz}{{\bf z}}

\newcommand{\vA}{{\bf A}}
\newcommand{\vB}{{\bf B}}
\newcommand{\vC}{{\bf C}}

\newcommand{\vI}{{\bf I}}

\usepackage{booktabs, multirow, bbding, bm, amssymb}

\usepackage[listings]{tcolorbox}
\definecolor{Gray}{gray}{0.94}

\usepackage[inline, shortlabels]{enumitem}